\def\year{2022}\relax
\pdfoutput=1 
\documentclass[letterpaper]{article} 
\usepackage{aaai22}  
\usepackage{times}  
\usepackage{helvet}  
\usepackage{courier}  
\usepackage[hyphens]{url}  
\usepackage{graphicx} 
\usepackage{natbib}  
\usepackage{caption} 
\DeclareCaptionStyle{ruled}{labelfont=normalfont,labelsep=colon,strut=off} 
\usepackage{algorithm}
\usepackage{algorithmic}

\usepackage{newfloat}
\usepackage{listings}
\floatstyle{ruled}
\newfloat{listing}{tb}{lst}{}
\floatname{listing}{Listing}
\usepackage{amsmath}        
\usepackage{amsfonts}       
\usepackage{nicefrac}       
\usepackage{bm}             
\usepackage[caption=false]{subfig}

\usepackage{booktabs}  

\usepackage{chapterbib}     

\usepackage{tikz}           
\usetikzlibrary{automata, positioning}
\usetikzlibrary{arrows, arrows.meta}

\usepackage{pgf}            
\usepackage{scalefnt}       

\usepackage[page]{appendix} 

\newcommand{\Comment}[1]{\hfill$\triangleright$ {#1}}

\usepackage{amsthm}
\newtheorem{theorem}{Theorem}
\newtheorem{proposition}{Proposition}

\newtheorem{lemma}[theorem]{Lemma}

\newcommand{\vect}[1]{\boldsymbol{\mathbf{#1}}}
\newcommand{\vn}[1]{\vect{#1}}

\newcommand{\vZ}{\vect{\Xi}}
\newcommand{\vphi}{\vect{\phi}}
\newcommand{\vPhi}{\vect{\Phi}}
\newcommand{\vtheta}{\vect{\theta}}
\newcommand{\vpsi}{\vect{\psi}}
\newcommand{\vw}{\vect{w}}

\newcommand{\vflet}{{\eta}}  
\newcommand{\returnname}{{$\eta$-return mixture}}

\title{A Generalized Bootstrap Target for Value-Learning, \\ Efficiently Combining Value and Feature Predictions}

\author{
    Anthony GX-Chen \textsuperscript{\rm 1}\textsuperscript{$\dagger$}
    Veronica Chelu \textsuperscript{\rm 1},
    Blake A. Richards \textsuperscript{\rm 1} \textsuperscript{\rm 2},
    Joelle Pineau \textsuperscript{\rm 1} \textsuperscript{\rm 2} \textsuperscript{\rm 3}
}
\affiliations{
    \textsuperscript{\rm 1}Mila - Quebec AI Institute \& McGill University \\
    \textsuperscript{\rm 2}Canada AI Chair program, CIFAR, Learning in Machines \& Brains \\
    \textsuperscript{\rm 3}Meta AI Research \\

    \textsuperscript{$\dagger$}{\rm anthony.gx.chen@gmail.com}
}

\begin{document}

\maketitle

\begin{abstract}
    Estimating value functions is a core component of reinforcement learning algorithms. Temporal difference (TD) learning algorithms use \emph{bootstrapping}, i.e. they update the value function toward a learning target using value estimates at subsequent time-steps. Alternatively, the value function can be updated toward a learning target constructed by separately predicting successor features (SF)---a policy-dependent \emph{model}---and linearly combining them with instantaneous rewards.
    We focus on bootstrapping targets used when estimating value functions, and propose a new backup target, the \emph{\returnname}, which implicitly \emph{combines} value-predictive knowledge (used by TD methods) with (successor) feature-predictive knowledge---with a parameter $\vflet$ capturing how much to rely on each. We illustrate that incorporating predictive knowledge through an $\vflet\gamma$-discounted SF model makes more efficient use of sampled experience, compared to either extreme, i.e. bootstrapping entirely on the value function estimate, or bootstrapping on the product of separately estimated successor features and instantaneous reward models. We empirically show this approach leads to faster policy evaluation and better control performance, for tabular and nonlinear function approximations, indicating scalability and generality.
\end{abstract}

\noindent
The fundamental goal of reinforcement learning (RL) is to maximize return, i.e. (temporally discounted) cumulative reward. Value functions provide an estimate of the expected return from a specific state (and action), and as such, they are a fundamental component of RL algorithms. Modern deep RL methods require numerous environment interactions to solve complex tasks, which can be expensive or impossible to obtain, particularly for tasks resembling the real-world. This makes it essential to develop data-efficient methods for learning accurate value functions.

The problem we address in this work is that of credit assignment, namely how to associate (distant) rewards to the states and actions that caused them. Value-based RL methods tackle this problem through \emph{temporal difference} (TD) learning algorithms \citep{Sutton1988LearningTP}. TD algorithms rely on \emph{bootstrapping}: using the \emph{value estimate} at a subsequent timestep, together with the observed data (e.g. rewards), to construct the learning \emph{target}---the \emph{return}---for the current timestep.
However, the value estimate in the backup target does not need to come from the current value function being learned. For instance, value can be estimated using \textit{successor features}---the (discounted) cumulative features---linearly combined with an estimate of instantaneous rewards \citep{barreto2017successor}. This approach can make use of the same TD methods \citep{Sutton1988LearningTP} to estimate the successor features as the former does when learning the value function, requiring similar amounts of sampled experience. Moreover, the \emph{backup target} and the value function can be completely distinct (e.g. if the successor features and learned value function are dis-jointly parameterized); they can share feature representations (e.g. when the value function and the successor features are both linear functions of the features); or partially share representations (e.g. through Polyak averaging). Since the value function is regressed toward the target, the method of computing the target influences the quality of the value function.


In this paper, we aim to improve credit assignment and data efficiency for value-based methods, by proposing a new method of constructing a learning target, which borrows properties from all aforementioned approaches of target construction. 
This \textbf{\returnname} uses a parameter $\vflet$ to combine an $\vflet\gamma$-discounted successor features model ($\vflet\gamma$-SF) \emph{with} the current value function estimate to parameterize the learning target used during bootstrapping---with the $\vflet$ parameter controlling the combination of value-predictive and feature-predictive knowledge.
We observe an intermediate value of $\vflet$ incorporates the benefits of both approaches in a complementary way, using sampled experience more efficiently.


\textbf{Contributions} In this paper we make three contributions: (i) We introduce the \emph{\returnname}, a simple yet novel way of constructing a backup target for value learning, using an $\vflet\gamma$-discounted SF model to interpolate between a direct value estimate and the fully factorized estimate relying on SF and instantaneous rewards. (ii) We describe a new learning algorithm using the \emph{\returnname} as the bootstrap target for value estimation. (iii) We provide empirical results showing more efficient use of experience with the \emph{\returnname} as the backup target, in both prediction and control, for tabular and nonlinear approximation, when compared to baselines.

\section{Preliminaries}
We denote random variables with uppercase (e.g., $S$) and the obtained values with lowercase letters (e.g., $S \!=\! s$). Multi-dimensional functions or vectors are bolded (e.g., $\mathbf{w}$), as are matrices (e.g. $\mathbf{\Phi}$). For all state-dependent functions, we also allow time-dependent shorthands (e.g., ${\vphi}_t\!=\! \vect{\phi}(S_t)$).

\subsection{Reinforcement learning problem setup}
A discounted Markov Decision Process (MDP) \citep{puterman1994mdp} is defined as the tuple $(\mathcal{S}, \mathcal{A}, P, r)$, with state space $\mathcal{S}$, action space $\mathcal{A}$, reward function $r: \mathcal{S} \times \mathcal{A} \to \mathbb{R}$, and transition probability function $P:\mathcal{S}\times\mathcal{A} \times\mathcal{S}\to \mathcal{P}(\mathcal{S})$ (with $\mathcal{P}(\mathcal{S})$ the set of probability distributions on $\mathcal{S}$, and $P(s^\prime|s,a)$ the probability of transitioning to state $s^\prime$ by choosing action $a$ at state $s$). A policy $\pi: \mathcal{S} \to \mathcal{P}(\mathcal{A})$ maps states to distributions over actions; $\pi(a|s)$ denotes the probability of choosing action $a$ in state $s$. Let $S_t, A_t, R_t$ denote the random variables of state, action and reward at time $t$, respectively.

Policy evaluation implies estimating the value function $v_\pi$, defined as the expected discounted return:
\begin{align}
    G_t &\equiv
        R_{t+1} + \textstyle\sum\limits _{k=1}^{\infty}\gamma^{k} R_{t+k+1} =  R_{t+1} + \gamma G_{t+1} \,,
    \\
    v_\pi(s) &\equiv \mathbb{E} \left[G_t \mid S_t = s, A_k \sim \pi(S_k), k \geq t
    \right]\,,
    \label{eq:def-value}
\end{align}
where $\gamma \in [0, 1)$ is the discount factor.
The learner's goal is to find a policy, $\pi$ which maximizes the \textit{value} $v_\pi$. When the Markov chain induced by $\pi$ is ergodic, we denote with $d_\pi$ the stationary distribution induced by policy $\pi$.
We henceforth shorthand the expectation over the environment dynamics and the policy $\pi$ with $\mathbb{E}_\pi[\cdot]$.

\subsection{Value learning}

Typically, $v_\pi$ is represented directly, using a linear parametrization over some state features $\vphi(s) \in \mathbb{R}^d$, where $d$ is the dimension of the representation space:
\begin{equation}
    v_{\vtheta} (s) = \vect{\phi(s)}^\top \vtheta  \approx v_\pi(s) ,
    \label{eq:model-free-value-function}
\end{equation}
with $\vtheta \in \mathbb{R}^d$ learnable parameters, and $\vphi(s)$ are features.\footnote{$\vphi(s)$ can be a parameterized non-linear function jointly learned with $\vtheta$, as is the case for many end-to-end deep reinforcement learning algorithms.} Learning $v_\pi$ with TD methods involves bootstrapping on a \textit{target}, $U_t$, at each timestep $t$, and updating $\vtheta$ by regressing it towards the target:
\begin{equation}
    \vtheta^\prime = \vtheta + \alpha \left[U_t - v_{\vtheta}(S_t)\right]\nabla\!_{\vtheta} v_{\vtheta}(S_t)\,,
    \label{eq:value-parameter-regression-update}
\end{equation}
with learning rate $\alpha$. The TD($0$) algorithm \citep{Sutton1988LearningTP} uses the \textit{one-step TD return} as the value target:
\begin{equation}
    U_t \equiv G^{(0)}_t = R_{t+1} + \gamma v_{\vtheta}(S_{t+1})\,.
    \label{eq:one-step-td-return}
\end{equation}
The forward view of TD($\lambda$) constructs the \emph{$\lambda$-return} target---a geometrically weighted average over all possible multi-step returns (\citet{sutton2018reinforcement}, chapter 12.1):
\begin{align}
    U_t \equiv G_t^\lambda &=\! (1-\lambda) \textstyle\sum\limits_{n=1}^\infty \lambda^{n-1} G_t^{(n)} \,, \text{ with}\label{eq:lambda-return-full}
    \\
    G_t^{(n)}&\equiv \big( \textstyle\sum\limits_{k=1}^{n} \gamma^{k-1} R_{t+k} \, \big) + \gamma^n v_{\vtheta}(S_{t+n}) \,,
\end{align}
where $\lambda \in [0,1]$ controls the weight of value estimates from the distant future, interpolating between the one-step return (equation~\eqref{eq:one-step-td-return}) ($\lambda = 0$) and the Monte Carlo return ($\lambda = 1$). The $\lambda$-return can only be computed offline at the end of an episode, since it requires the entire future trajectory to calculate the multi-step returns.

\subsection{Successor features (SF)}
Previous work \citep{dayan1993improving,kulkarni2016deepsuc,zhang2017deep,barreto2017successor,barreto2018transfer} has shown it can be useful to decouple the reward and transition information of the value function by factorizing it into immediate rewards and
SF. The SF, $\vpsi_{\pi} : \mathbb{R}^d \to \mathbb{R}^d$, are defined as the expected cumulative discounted features under a policy $\pi$:
\begin{align}
    {\vpsi}_{\pi}(s) \!\equiv\! \mathbb{E}_\pi \left[
        \textstyle\sum_{n\!=\!0}^{\infty} \gamma^n {\vphi}_{t\!+\!n}
   \! \mid\! S_t\!=\!s\right]\,,
   \label{eq:full-sf-def}
\end{align}
and can be learned by TD learning algorithms, similar to the standard value function:
\begin{align}
   {\vpsi}_{\vZ}(s) &=\vZ^\top\vphi(s)\approx \vpsi_\pi(s) \,,\text{with}
   \label{eq:sf-model}
   \\
   \vZ^\prime &= \vZ + \alpha \, \delta_{\vZ} \, \nabla\!_{\vZ} \vpsi_{\vZ}(S_t)\,
   \label{eq:sf-parameter-update}
   \\
   \delta_{\vZ} &\equiv \vphi(S_t) + \gamma \vpsi_{\vZ}(S_{t\!+\!1}) - \vpsi_{\vZ}(S_{t})  \,,
\end{align}
with ${\vZ} \in\mathbb{R}^{d \times d}$ (learnable) parameters.\footnote{Unless stated otherwise, we consider $\vpsi$ as a linear function of features (equation~\eqref{eq:sf-model}), though non-linear functions are available
\citep{zhang2017deep,machado2020count}.}
An alternative approach to the direct representation of value (equation (\ref{eq:model-free-value-function})) is to used a factorization of SF and instantaneous reward:
\begin{align}
    v_{\vpsi} (s) &\equiv \vpsi_{\vZ}(s)^\top \textbf{w} \approx v_\pi(s)\,,  \text{ with }
    \label{eq:full-sf-value-function}
    \\
    r_{\textbf{w}}(s) &\equiv \vphi(s)^\top \textbf{w} \approx \mathbb{E}_\pi [R_{t+1} | S_t = s] \,,
    \label{eq:linear-reward-function}
\end{align}
the instantaneous reward function with (learnable) parameters $\textbf{w} \in \mathbb{R}^d$.

\section{The \returnname}

We take inspiration from the canonical $\lambda$-return (equation~\eqref{eq:lambda-return-full}), to write a similar quantity.\footnote{We replaced $\lambda$ with $\vflet$ to denote the  different properties of the interpolation parameter $\vflet$ compared to $\lambda$.} A full derivation of this section is given in appendix \ref{Ap:lambda-return-equivalencies}.
\begin{align}
    G_t^{\vflet} \approx R_{t+1} + \gamma \Big[
        (1-\vflet) & \textstyle\sum\limits_{n=1}^\infty (\vflet \gamma)^{n-1} v_{\vtheta} (S_{t+n}) \nonumber
        \\ + \, & \vflet  \textstyle\sum\limits_{n=1}^\infty (\vflet \gamma)^{n-1} r_{\vw} (S_{t+n})
    \Big] \,.
    \label{eq:eta-return-summation-def}
\end{align}
As both $v_{\vtheta}$ (equation (\ref{eq:model-free-value-function})) and $r_{\vw}$ (equation (\ref{eq:linear-reward-function})) are linear in features, we can express the geometric sums in equation (\ref{eq:eta-return-summation-def}) using $\vflet\gamma$-discounted SFs,
\begin{align}
    \vpsi^{\vflet}(s) &\equiv \mathbb{E}_\pi\!\left[
        \textstyle\sum_{n=0}^{\infty} (\vflet \gamma)^n \vphi_{t+n} \mid S_t=s \right]\,.
    \label{eq:etalambda-discounted-sf-def}
\end{align}
We can separately estimate this SF-model using equation~\eqref{eq:sf-parameter-update}. Further, we can use the SF-model in the bootstrapping process by substituting equation (\ref{eq:etalambda-discounted-sf-def}) into equation (\ref{eq:eta-return-summation-def}). This yields a learning target which uses predictive features ($\vpsi^\vflet$), along with a \emph{mixture} of value ($\vtheta$) and reward ($\vw$) parameters. This is the \textbf{\returnname}:
\begin{align}
    U_t \equiv G^\vflet_t &\equiv
        R_{t+1} + \gamma \vpsi^\vflet(S_{t+1})\left[(1-\vflet) \vtheta + \vflet \vw\right]\,.
    \label{eq:eta-return-mixture-target-def}
\end{align}
This target can be used to replace e.g. the standard TD($0$) backup target from equation~\eqref{eq:one-step-td-return}.
Despite its similarity to the standard $\lambda$-return, the \returnname\, does not assume access to a full episodic trajectory.

\paragraph{Interpretation}{
Consider learning using single-step transition tuple $(S_t, A_t, R_{t+1}, S_{t+1})$. TD(0) propagates information locally from $S_{t+1}$ to $S_t$ by constructing a \emph{bootstrapping target}. Using the value function in the target (equation~\eqref{eq:one-step-td-return}) propagates only \emph{value information}; bootstrapping using the product of estimated SF and instantaneous rewards (equation~\eqref{eq:full-sf-value-function}) relies on separately learning the SF, which also uses TD($0$), and thus propagates only \emph{feature information}. We can more effectively use the same single-step of experience if we simultaneously use the sampled information to predict both the value \emph{and} the features, and update the value function using a mixture of both in the way specified in equation~\eqref{eq:eta-return-mixture-target-def}.

}

\paragraph{Fixed-point solution}{
With accurate SF and instantaneous reward models, one-step value-learning with the \returnname\, as bootstrapping target has the same fixed-point solution as the standard TD($0$) target, per the following.
\begin{proposition}
    Assume the SF parameters ${\vZ}$ have converged to their fixed-point solution, ${\vZ}_{\text{TD}(0)} = \mathbb{E}_{d_\pi}\![\vphi_t(\vphi_t - \vflet\gamma \vphi_{t+1})^{\top}]^{-1}\mathbb{E}_{d_\pi}\![\vphi_t \vphi_t^\top]$, and the instantaneous reward parameters have achieved the optimal solution $\mathbf{w} = \mathbb{E}_{d_\pi}\![\vphi_t \vphi_{t}^{\top}]^{-1} \mathbb{E}_{d_\pi}\![\vphi_t R_{t+1}]$, where $\mathbb{E}_{d_\pi}\![\cdot]$ denotes the expectation over the stationary distribution $d_\pi$ for policy $\pi$, which we assume exists under mild conditions \citep{tsitsiklis1997analysis}.
    Then, value learning using the \returnname as the target has the TD($0$) fixed point solution:
    \begin{equation}
       \vtheta_\vflet^* = \mathbb{E}_{d_\pi}\!\left[\vphi_t(\vphi_t - \gamma \vphi_{t+1})^{\top}\right]^{-1} \mathbb{E}_{d_\pi}\!\left[\vphi_t R_{t+1}\right] =  \vtheta_{\text{TD}(0)}.
    \end{equation}
    \label{prop: theta-fixed-point}
\end{proposition}
\begin{proof}
    In Appendix \ref{Ap:proof-of-fixed-point}. Follows from the linearity of the policy evaluation equations.
\end{proof}
Furthermore, it has been shown that on-policy planning with linear models converges to the same fixed point as direct linear value estimation \citep{Schoknecht2002OptimalityOR, Parr2008AnAO, Sutton2008DynaStylePW}. However, despite the fact that the fixed point solution is subject to the same bias as one-step TD methods, our method may still benefit from substantial learning efficiency while moving towards this solution. In fact, our finite sample empirical evaluation shows exactly this.
}

\begin{figure*}[ht]
    \centering

    \textbf{(A)} \hspace{10pt}
    \subfloat{\scalebox{0.52}{

\begin{tikzpicture}[scale=1]
    \tikzset{node style/.style={
        state, fill=white, circle}}
    \tikzset{start style/.style={
        state, circle,
        fill={violet!30!white}},
    }
    \tikzset{term style/.style={state, 
                                fill=gray!5!black,
                                rectangle}}
    \node[start style]              (1)  {1};
    \node[node style, right=of 1]   (2)  {2};
    \node[draw=none,  right=of 2]   (2-3) {$\cdots$};
    \node[node style, right=of 2-3] (3) {8};
    \node[draw=none,  right=of 3]   (3-4)  {$\cdots$};
    \node[node style, right=of 3-4] (4)   {15};
    \node[node style, right=of 4]   (5)   {16};
    \node[term style, right=of 5, scale=0.5] (f)   {f};

    \draw[>=latex,
            line width=1.6pt,
            auto=left,
            every loop]
        (1)     edge node   {}          (2)
        (2)     edge node   {}          (2-3)
        (2-3)   edge node   {}          (3)
        (3)     edge node   {}          (3-4)
        (3-4)   edge node   {}          (4)
        (4)     edge node   {}          (5)
        (5)     edge node   {$+1$}      (f);
\end{tikzpicture}

        }}  \hspace{110pt}

    \textbf{(B)} \hspace{-7pt}
    \subfloat{\scalebox{0.52}{
        \includegraphics{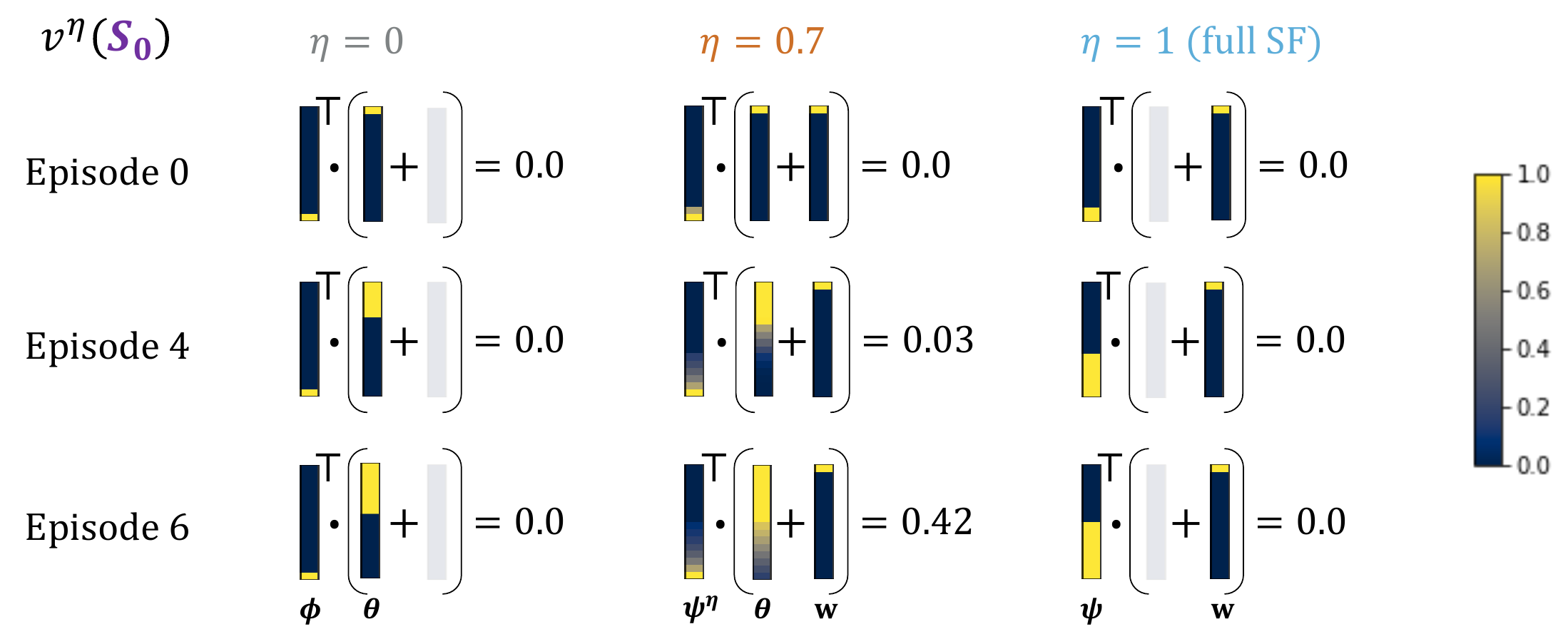}
        }} \vspace{5pt}

    \textbf{(C)} \hspace{-10pt}
    \subfloat{\scalebox{0.45}{
        \includegraphics{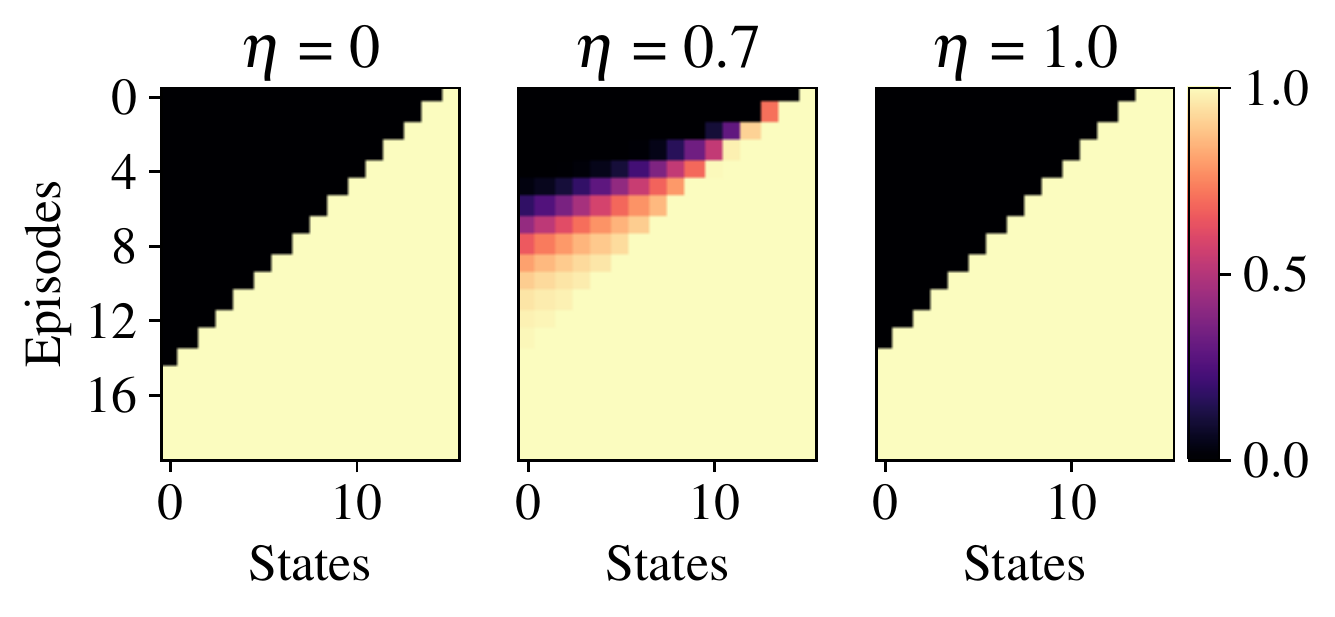}
        }} \hspace{20pt}
    \textbf{(D)} \hspace{-20pt}
    \subfloat{\scalebox{0.42}{
        \scalefont{10.0}
        \input{figs/intuit/det-chain_errors.pgf}
        }}

    \caption{\textbf{Online value prediction in a deterministic MRP for different $\vflet$'s.}. \textbf{(A)} The agent starts in the left-most state ($s_0$) and deterministically transitions right until reaching the terminal state. All rewards are $0$, except for the transition into the terminal state when it is $+1$. \textbf{(B)} \textbf{Parameter dynamics:} The table shows how the \returnname\,value estimate for the first state, $v^\vflet(S_0)$, is computed using ${\vpsi}^\vflet(s_0)^\top$, $\vtheta$ and $\vw$ over the course of training, for different values of $\vflet = \{0.0, 0.7, 1.0\}$. For $\vflet=0.7$ (center) the estimation for the \returnname\,combines the parameters of the value function ($\vtheta$) and the SF predictions ($\vpsi^\vflet$) to more quickly propagate value information than either extremes. \textbf{(C)} \textbf{The estimated value function} for all states (columns) across learning episodes (rows). For $\vflet=0.7$ value information propagates faster than $\vflet=0$ and $1$. \textbf{(D)} \textbf{Absolute value error:} for different $\vflet$ values over episodes. For $\vflet = 0.7$ error reduction is faster.}
    \label{fig:deterministic-chain-intuition}
\end{figure*}

\paragraph{Interpolating between value and feature prediction with $\vflet$}{
Similar to how $\lambda$-return interpolates between the one-step TD and Monte-Carlo returns, the \returnname\, interpolates between bootstrapping on the ``value-predictive'' parameters of the value function,  or  on the ``feature-predictive'' parameters of the SF.

When $\vflet=0$, the \returnname\, recovers the standard TD(0) learning target (equation~\eqref{eq:one-step-td-return}):
\begin{align}
    G_t^{\vflet=0} &= R_{t+1} + \gamma \vpsi^{\vflet=0}(S_{t+1})^\top \left((1-0) \vtheta + 0 \vw\right)\nonumber \\
    &= R_{t+1} + \gamma\vphi_{t+1}^\top \vtheta \,.
\end{align}

At the opposite end of the spectrum, when $\vflet=1$, the \returnname\, relies on the full SF (equation~\eqref{eq:full-sf-value-function}) and the instantaneous reward model, akin to using an implicit infinite model:
\begin{align}
    G_t^{\vflet=1} &= R_{t+1} + \gamma \vpsi^{\vflet=1} (S_{t+1})^\top \left((1-1) \vtheta + 1 \vw\right) \nonumber \\
    &= R_{t+1} + \gamma {\vpsi}_{t+1}^\top \vw \,.
\end{align}
Consequently, the \returnname\, is a simple generalization that spans the spectrum of learning target parameterizations using $\vflet \in [0, 1]$, with the traditional learning target and the SF factorization as extremes.

Compared to the standard learning target used in TD($0$), \textbf{the \returnname\, with an intermediate value of $\vflet$ ($0 < \vflet < 1$) uses information more effectively} than the extremes $G_t^{\vflet=0}$ (equation~\eqref{eq:one-step-td-return}), and $G_t^{\vflet=1}$, approximating the true value faster given the same amount of data (see figure~\ref{fig:deterministic-chain-intuition} for an intuitive illustration).
}

\subsection{Estimating the \returnname}

There are different choices with respect to how the learning target is estimated, depending on (i) the form or elements used in building the target; (ii) the parametrization of the elements making up the target; (iii) the learning methods used to estimate the elements of the target.

Regarding (i), the form of the \returnname\, target requires access to SF, instantaneous rewards, and the value parameters themselves. Regarding (ii), we parameterize all these estimators as linear functions of features, and share feature parameters in cases where the feature representation is learned and not given (e.g. in the nonlinear control empirical experiments).

With respect to (iii), we can use any learning method for estimating the SF model $\vpsi^{\vflet}_{\vZ}$ and the instantaneous reward model $r_\mathbf{w}$.
In this paper, we make the choice of using TD($0$) to learn the SF model, and supervised regression for the reward model, since one-step methods are ubiquitous in contemporary RL, and require the use of only \emph{single-step} transitions \citep{mnih2015human,van2015double,lillicrap2015ddpg,wang2016dueling,schaul2015prioritized,haarnoja2018sac}.
Likewise, we use the \returnname\, as a one-step bootstrap target (equation~\eqref{eq:eta-return-mixture-target-def}) for estimating of the value parameters $\vtheta$ (equation~\eqref{eq:value-parameter-regression-update}). Although we have chosen to focus here on one-step learning targets for their simplicity and ease of use, these methods can be extended to multi-step targets (e.g. TD($n$) or TD($\lambda$)) analogously as the one-step target.

\begin{algorithm}[ht!]
    \caption{Value prediction using a linear \returnname}
    \label{alg:linear-lambda-vf-prediction}
    \textbf{Input}:
    Given $v_{\vtheta}(s) = \vphi(s)^\top {\vtheta}$ (value function),
    $r_{\vw}(s) = \vphi(s)^\top {\vw}$ (instantaneous reward function),  $\vpsi_{\vZ}^\vflet(s) = {\vZ}^\top \vphi(s)$ (SF model),
    $\gamma \in [0, 1)$, $\vflet \in [0,1]$,
    $\alpha^{\vtheta}$, $\alpha^{\vw}$, $\alpha^{\vZ}$ (learning rates). \\
    \textbf{Output}: Value function ${v}_{\vtheta}\approx v_\pi$ for a policy $\pi$. \\
    \vspace{-1em} 
    \begin{algorithmic}[1] 
        \WHILE{sample one-step experience tuple using $\pi$, $(S_t, A_t, R_{t+1}, S_{t+1})$,}
            \STATE SF learning update: ${\vZ}_{t+1} \leftarrow {\vZ}_{t} + \alpha^{\vZ}_t (
            \vphi(S_t) + \vflet\gamma \vpsi_{{\vZ}_t}^\vflet(S_{t+1}) - {{\vZ}_t}^\top \vphi(S_t)
            )\, \vphi(S_t)^\top $

            \STATE Instantaneous reward learning update: \\ ${\vw}_{t+1} \leftarrow {\vw}_t \!+\! \alpha^{\vw}_t (
                R_{t+1} \!- \!\vphi(S_t)^\top {{\vw}}_t
            ) \, \vphi(s)$

            \STATE Next step value estimate: \\ $v^\vflet_{t+1} = \vphi(S_{t+1})^\top {\vZ}_{t+1} ( (1\!-\!\vflet) {\vtheta}_t \!+\! \vflet {\vw}_{t+1} )$

            \STATE Value learning update: \\ ${\vtheta}_{t+1} \leftarrow {\vtheta}_t \!+\! \alpha^{\vtheta}_t (
                R_{t+1} \!+ \!\gamma v^\vflet_{t+1}\! -\! \vphi(S_t)^\top {\vtheta}_t
            ) \vphi(S_t)$
        \ENDWHILE
    \end{algorithmic}
\end{algorithm}

All components of the \returnname\, are now learnable with one-step transitions tuples of the form $(S_t, A_t, R_{t+1}, S_{t+1})$, which make these methods amenable to both the online setting and the i.i.d. setting. In the former, the algorithm is presented with an infinite sequence of state, actions, rewards $\{S_0, A_0, R_1, S_1, A_1, R_2, \dots\}$, where $A_t \sim \pi(S_t), R_{t+1} \equiv r(S_t, A_t), S_{t+1} \sim P(S_t, A_t)$. In the i.i.d. setting, the learner is presented with a set of transition tuples $\{(S_t, A_t, R_{t+1}, S_{t+1})\}_{t\geq0}$.

From an algorithmic perspective, we now describe a computationally congenial way for learning the value function online, from a single stream of experience, using our method. As mentioned, in the online setting, the agent has access to experience in the form of tuples $(S_t, A_t, R_{t+1}, S_{t+1})$ at each timestep $t$. The pseudo-code in algorithm~\ref{alg:linear-lambda-vf-prediction} describes the online value estimation process, for the linear case, with given representations.

\section{Empirical studies}
\label{sec:empirical-experiments}

We start with two simple prediction examples to provide intuition about our approach, after which, we verify that our method scales by extending it to a more complex non-linear control setting.

\begin{figure*}[ht]
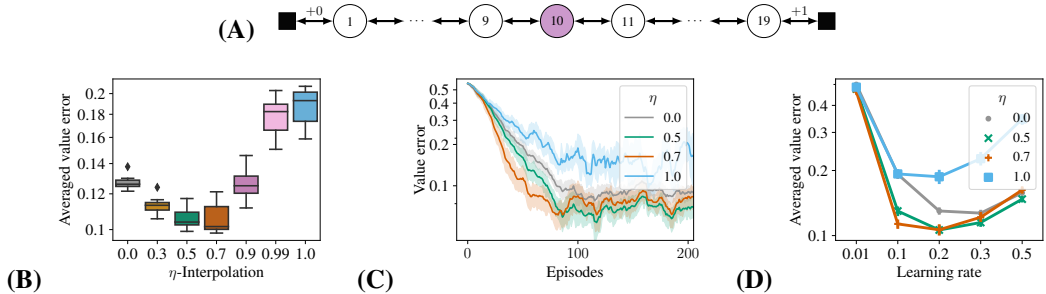

    \centering

    \textbf{(A) \;}
    \subfloat{\scalebox{0.5}{

\begin{tikzpicture}[scale=1]

    \tikzset{node style/.style={state, 
                                fill=white,
                                circle}}
    \tikzset{start style/.style={
        state, circle,
        fill={violet!40!white}},
    }
    \tikzset{term style/.style={state, 
                                fill=gray!5!black,
                                rectangle}}
                                
    \node[term style, scale=0.5]    (0)     {};
    \node[node style, right=of 0]   (1)     {1};
    \node[draw=none, right=of 1]   (1-9)   {$\cdots$};
    \node[node style, right=of 1-9]   (9)     {9};
    \node[start style, right=of 9]   (10)    {10};
    \node[node style, right=of 10]   (11)    {11};
    \node[draw=none, right=of 11]   (11-19) {$\cdots$};
    \node[node style, right=of 11-19]   (19)    {19};
    \node[term style, right=of 19, scale=0.5]   (20)    {};
    
    \draw[latex-latex, line width=1.6pt] (0) -- node[above] {$+0$} (1);
    \draw[latex-latex, line width=1.6pt] (1) -- node[] {} (1-9);
    \draw[latex-latex, line width=1.6pt] (1-9) -- node[] {} (9);
    \draw[latex-latex, line width=1.6pt] (9) -- node[] {} (10);
    \draw[latex-latex, line width=1.6pt] (10) -- node[] {} (11);
    \draw[latex-latex, line width=1.6pt] (11) -- node[] {} (11-19);
    \draw[latex-latex, line width=1.6pt] (11-19) -- node[] {} (19);
    \draw[latex-latex, line width=1.6pt] (19) -- node[above] {$+1$} (20);

\end{tikzpicture}

        }}

    \textbf{(B)}
    \subfloat{\scalebox{0.4}{
        \large
        \input{figs/rchain/random-chain_best-per-lambda.pgf}
        }}
        \quad
    \textbf{(C)}
    \subfloat{\scalebox{0.4}{
        \input{figs/rchain/random-chain_learning-dynamics.pgf}
        }}
        \quad
    \textbf{(D)}
    \subfloat{\scalebox{0.4}{
        \input{figs/rchain/random-chain_lr-study.pgf}
        }}
    \caption{\textbf{Policy evaluation in 19-state tabular random chain.} \textbf{(A)} The agent starts in the center and transitions left/right randomly until either end is reached. Reward is $0$ on all transitions, except the on the right-side termination, which yields a reward of $+1$.  \textbf{(B)} \textbf{Parameter study for $\vflet$:} The y-axis shows the root mean squared error (RMSE) (minimized over learning rates for each $\vflet$) averaged over first $400$ episodes. \textbf{(C)} \textbf{Learning dynamics:} The y-axis shows the RMSE for four illustrative $\vflet$ values. \textbf{(D)} \textbf{Parameter study for the learning rate} The y-axis shows the RMSE for four illustrative $\vflet$ values, across different learning rates. Results averages over first $400$ episodes. Error bars and shaded areas denote $95$ confidence intervals (some too small to see), with $10$ independent seeds.}
    \label{fig:tabular-random-chain}
\end{figure*}

\subsection{Value prediction in a deterministic chain}
\label{sec:deterministic-chain-value-prediction}

\textbf{Experiment setup:}
Consider the 16-state deterministic Markov reward process (MRP) with tabular features illustrated in figure \ref{fig:deterministic-chain-intuition}-A. The agent starts in the left-most state ($s_0$), deterministically transitions right to the right-most absorbing state. The reward is $0$ everywhere except for the final transition into the absorbing state, where it is $+1$. We apply algorithm~\ref{alg:linear-lambda-vf-prediction} to estimate the value function in an online incremental setting. We use a discount factor $\gamma=0.9999$ and learning rate $\alpha=1.0$.

\noindent \textbf{Results: }
Figure \ref{fig:deterministic-chain-intuition}-B illustrates the result of combining the successor features model $\vpsi_{\vZ}$, with the value parameters $\vtheta$, and reward parameters $\vw$ into a prediction of the \returnname\, for the starting state $s_0$, $v^{\vflet} (s_0)$, for different values of $\vflet$. When completely relying on the canonical value bootstrap target ($\vflet=0$, recovering TD(0)), we have $v^{\vflet=0} = v_{\vtheta}$, which corresponds to an unchanging feature representation. In this setting, the value information (in $\vtheta$) moves \textit{backward} one state per episode. For the opposite end, when bootstrapping on the full successor features (${\vflet}=1$), the instantaneous reward is learned immediately (parameter $\vw$) for the final state, while the successor features (parameter $\vpsi$) learns about one additional future state per episode. For both cases, we require $\sim 16$ episodes for the information to propagate across the entire chain and for the value estimate of $s_0$ to improve (Figure \ref{fig:deterministic-chain-intuition}-D). However, with an intermediate value of $0 < {\vflet} < 1$ (Figure \ref{fig:deterministic-chain-intuition}-B, middle, ${\vflet}=0.7$ here), we are able to both propagate value information backward by bootstrapping on $\vtheta$, as well as improve the predictive features (using $\vpsi^{\vflet}$) to predict further in the forward direction. This results in an improved value estimate much earlier, as we can observe in figure~\ref{fig:deterministic-chain-intuition}-B middle, C  middle, and D.

\noindent\textbf{Interpretation:} In an online prediction setting, using the \returnname\, (with an intermediate $\vflet$: $0 < \vflet < 1$), in place of the standard TD($0$) learning target, effectively combines both \textit{backward} credit assignment by bootstrapping the value estimates, as well as \textit{forward} feature prediction, to more quickly estimate the correct values.

\subsection{Value prediction in a random chain}
\label{sec:random-chain-value-prediction}

\textbf{Experiment setup:}
We now switch to a slightly harder setting, a stochastic $19$-state chain prediction task, still with tabular features \citep[][Example 6.2]{sutton2018reinforcement}. The agent starts in the centre (state $10$) and randomly transitions left or right until reaching the absorbing states at either end (figure \ref{fig:tabular-random-chain}-A). The reward is $0$ everywhere except upon transitioning into the right-most terminal state, when it is $+1$. Hyperparameters were chosen by sweeping over learning rates $\alpha \in \{0.01, 0.1, 0.2, 0.3, 0.5\}$,  and mixing parameter $\vflet = \{0.0, 0.3, 0.5, 0.7, 0.9, 0.99, 1.0\}$. Figure  \ref{fig:tabular-random-chain}-B,D illustrate value error averaged over the first $400$ episodes.

\noindent\textbf{Results:}
In figure~\ref{fig:tabular-random-chain}-B, we observe that mixing with $\vflet \in [0, 1]$ results in a U-shape error curve, illustrating that an intermediate value of $\vflet$ is optimal. For each value of $\vflet$, we plot the optimal learning rate $\alpha$. Figure~\ref{fig:tabular-random-chain}-C further confirms our hypothesis that an intermediate value (here for $\vflet = 0.5$ or $0.7$) is most efficient. We also observe that intermediate $\vflet$'s show a degree of parameter robustness, having low value error over a range of different learning rates (figure~\ref{fig:tabular-random-chain}-D).

\noindent\textbf{Interpretation:}
Using the \returnname\,as one-step learning target is robust to environment stochasticity and learns most efficiently for intermediate values of $\vflet$.

\begin{figure*}[ht]
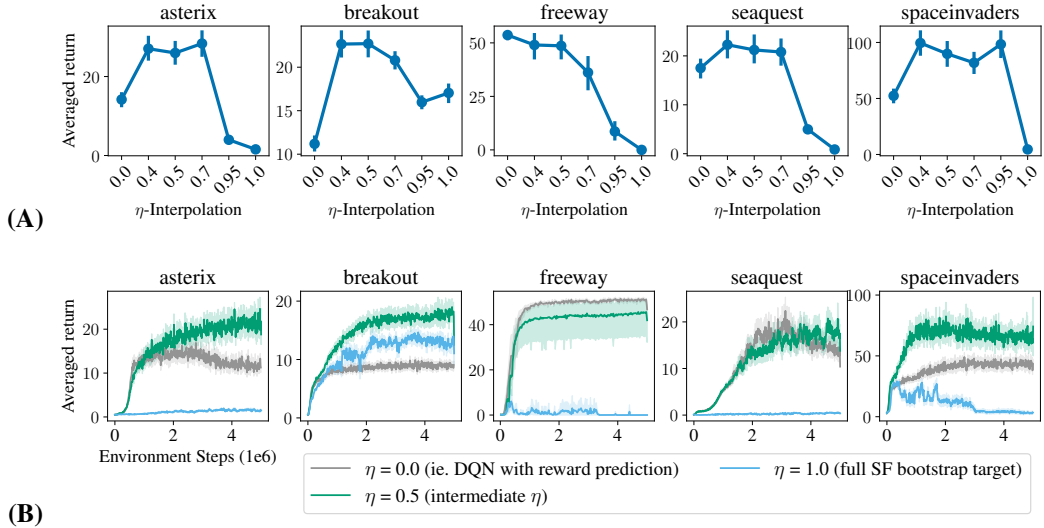

    \centering
    \textbf{(A)}
    \subfloat{\scalebox{0.45}{
        \input{figs/minatar/minatar_eval_lambdas.pgf}
    }} \vspace{0em}

    \textbf{(B)}
    \subfloat{\scalebox{0.45}{
        \input{figs/minatar/minatar_train_return.pgf}
    }}

    \caption{\textbf{Performance for value-based control in Mini-Atari}. \textbf{(A)} Parameter study for different values of $\vflet$. The y-axis shows the average performance over $10$k timesteps and $10$ seeds using an $\epsilon$-greedy policy with $\epsilon=0.05$, after stopping training after $5e6$ learning steps. \textbf{(B)} Learning curves for $3$ illustrative $\vflet$ values over the course of training. The y-axis displays the average return over $10$ independent seed. Shaded area and error bars depicts $95$ confidence interval.}
    \label{fig:minatar-q-performance-overview}
\end{figure*}

\subsection{Value-based control in Mini-Atari}
\label{sec:minatar-control}
We hypothesize that efficient value prediction using the $\vflet$-return can help in value-based control, so we extend our proposed algorithm to the control setting, simply by estimating the action-value function $q_{\vtheta}$ using the \returnname. We build on top of the deep Q network (DQN) architecture \citep{mnih2015human}, and simply replace the bootstrap target with an estimate of the \returnname\, starting from a \emph{state and action}.

Given a sampled transition $(S_t, A_t, R_{t+1}, S_{t+1})$, DQN encodes features $\vphi_t = \vphi(S_t)$, then estimates the action-values $q_{\vtheta} (\vphi_t, A_t) \approx q(S_t, A_t)$ using the canonical bootstrap target in which it relies on the next value estimate, $\max_{a^\prime} q_{\vtheta} (\vphi_{t+1}, a^\prime)$, with $\vphi_{t+1}=\vphi(S_{t+1})$.
We use the same feature encoding $\vphi(
\cdot)$ to track the successor features of the current policy $\vpsi^\vflet_t = \vpsi^{\vflet}_{\vZ} (\vphi_t) \approx \vpsi(\vphi_t)$, and estimate the instantaneous rewards $r_{\vw}(\vphi_t)$. This allows us to construct the \returnname\,and use it in the learning target of Q-learning when updating the parameters $\theta$:
\begin{align}
    &q^\vflet_{t+1} \equiv (1-\vflet)\, q_{\vtheta}\!\big({\vpsi}^\vflet_{\vZ}(S_{t+1}), a^\prime \big) + \vflet \, r_{\vw}\!\big({\vpsi}^\vflet_{\vZ}(S_{t+1})\big) \,,
    \\
    &\vtheta^{\prime} = \vtheta + \alpha \big( R_{t+1} + \gamma \max_{a^\prime} q^\vflet_{t+1} 
    - q_{\vtheta}({\vphi}_t, A_t) \big) \, \nabla\!_{\vtheta}q_{\vtheta}\,,
\end{align}
where $q^\vflet$ is the value estimate of the \returnname\,used in the learning target, and $\nabla\!_{\vtheta}q_{\vtheta} = \nabla\!_{\vtheta}q_{\vtheta}({\vphi}_t, A_t)$. We simultaneously estimate the feature representation and the action-values in an end-to-end fashion. See Appendix algorithm \ref{alg:deep-lambda-q-learning} for a complete description.

\noindent\textbf{Experiment Set-up: }
We test our algorithm in the Mini-Atari (MinAtar, \citet{young2019minatar}, GNU General Public License v3.0) environment, which is a smaller version of the Arcade Learning Environment \citep{bellemare2013arcade} with 5 games (\texttt{asterix, breakout, freeway, seaquest, space invaders}) played in the same way as their larger counterparts. Other than the architectural update to the bootstrap target, we make no other changes (e.g. to policy, relay buffer, etc.). Unless otherwise stated, we use the same hyperparameters as DQN version from \citet{young2019minatar}.  Details on environment, algorithms and hyperparameters can be found in appendix \ref{Ap:experiment-details}.

\textbf{Intermediate $\vflet$ improves nonlinear control.} Figure \ref{fig:minatar-q-performance-overview}-A illustrates a parameter study on the mixing parameter $\vflet$ after training for $5$ million environmental steps. We again observe the U-shaped performance curve as we interpolate across $\vflet$, confirming the advantage of using an intermediate $\vflet$ value. Figure \ref{fig:minatar-q-performance-overview}-B shows the learning curves of our proposed model that uses an intermediate value of $\vflet$ in comparison to the two baseline algorithms: bootstrapping entirely on the value parameters ($\vflet=0$, equivalent to vanilla DQN with a reward prediction auxiliary loss), and bootstrapping entirely on the full SF value ($\vflet=1$). The latter baseline is remarkably unstable, while the \returnname, with an intermediate $\vflet=0.5$, outperforms both in $\nicefrac{4}{5}$ games, and is competitive with $\vflet=0$ in  $\texttt{freeway}$. The poor performance for higher $\vflet$ values in $\texttt{freeway}$ is likely due to \emph{sparse reward}, as the reward gradient used to shape the representation $\vphi(\cdot)$ is uninformative most of the time, leading to a collapse in representation (this is explicitly measured in appendix section \ref{Ap:additional-results}). This highlights a weakness of learning the feature encoding and SF simultaneously, since poor features result in poor SF, and thus poor value estimates. The use of auxiliary losses can help ameliorate this issue \citep{machado2020count,kumar2020implicit}, although it is not explored here as we found the issue to only be significant for high values of $\vflet$.

\begin{figure*}[ht]
    \centering
    \subfloat{\scalebox{0.45}{
        \input{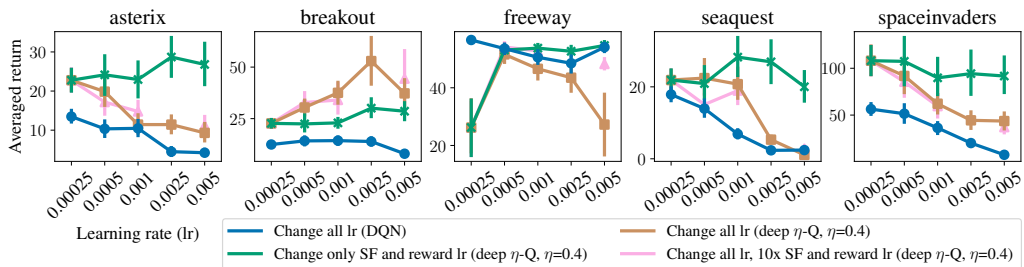}
    }}

    \caption{\textbf{Parameter study on the learning rates of the SF and instantaneous reward model:} The y-axis shows the average return over $10$k evaluation steps using an $\epsilon$-greedy policy with $\epsilon = 0.05$, after stopping training after $5e6$ steps. For our algorithm, shown here as the Deep $\vflet$-Q algorithm (green), we sweep over the SF and reward learning rates while keeping the learning rates for the representation torso and the value function head fixed at $0.00025$. For the vanilla DQN (blue), we vary the learning rates of the representation torso and the value function head. We also show the cumulative sensitivity to the parameters as we vary all the learning rates in our algorithm (brown). Error bar denote $95$ confidence intervals and each setting is ran using $3$ independent seeds.}
    \label{fig:minatar-lr-sweep}
\end{figure*}

\textbf{Parameter study: robustness to the learning rates of the SF and instantaneous reward models.} Figure~\ref{fig:minatar-lr-sweep} shows parameter studies for an intermediate $\vflet$ that illustrate the sensitivity to the learning rates of the successor features and reward heads used in learning the value function. We vary the learning rates for these estimators while keeping the learning rates of the representation torso and the value function head fixed (at the same values used by \citet{young2019minatar}: $\alpha_{\vtheta} = \text{2.5e-4}$). We observe that performance is not highly dependent on the SF and reward learning rates (figure~\ref{fig:minatar-lr-sweep}, green), but a higher learning rate for the SF than the one used by the representation torso facilitates tracking the changes in the feature representations ($\vphi$) by the SF. This choice is important in $\texttt{freeway}$. For comparison, we also sweep over the value and encoder learning rates of a vanilla DQN (figure~\ref{fig:minatar-lr-sweep}, blue), and see that it is sensitive to the learning rate, i.e. performance drops as learning rate settings deviate from the recommendation of \citet{young2019minatar} (most prominently observed in $\texttt{asterix}$, $\texttt{seaquest}$ and $\texttt{space\_invaders}$, and for high learning rates in $\texttt{breakout}$). Additionally, we also sweep over the learning rates of \textit{all} parameters making up the \returnname\, used as target for the q-function: either keeping all learning rates the same (figure~\ref{fig:minatar-lr-sweep}, brown) or setting the successor feature and reward learning rates to be $10\times$ the encoder learning rates (figure~\ref{fig:minatar-lr-sweep}, pink). Overall, we again observe that the agent is most sensitive to learning rates in the value head and encoder torso: performance decreases in all games other than \texttt{breakout}.

\section{Related work}

\textbf{Successor features} (SF, equation~\eqref{eq:full-sf-def}) are an extension to state-based successor representations \citep{dayan1993improving}, allowing feature-based value functions to be factorized using a separately parameterized policy-dependent transition model and an instantaneous reward model \citep{kulkarni2016deepsuc,lehnert2020successor}. A wide variety of uses have been proposed for the SF: aiding in exploration \citep{janz2019successor,machado2020count}, option discovery \citep{Machado2017ALF,Machado2018EigenoptionDT}, and transferring across multiple goals \citep{lehnert2017advantages,zhang2017deep,ma2020universal,brantley2021successor}, in particular through the generalized policy improvement framework \citep{barreto2017successor,barreto2018transfer,borsa2018universal,hansen2019var,grimm2019disentangled}. Our method adds to this repertoire, by using the SF inside the learning target in bootstrapping methods.

\textbf{Forward model-based planning} can facilitate efficient credit assignment. Among the algorithms that address this topic, are Dyna-style methods, which use explicit models to generate fictitious experience, that they then leverage to improve the value function \citep{Schoknecht2002OptimalityOR, Parr2008AnAO, Sutton2008DynaStylePW, yao2009multi}. Closest to our method is the work by \citet{yao2009dynak,yao2009multi} which learns an explicit $\lambda$-model and uses it to generate fictitious experience for $k$-step updates to the value function. Our work is different in that the model we use is an implicit model, used to construct a learning target.  Particularly, the SF here are akin to models for implicit planning, aiding in speeding up the value learning process within a \textit{single-task} setup. Furthermore, we extend our method to learned non-linear feature representations and combine it with batch learning algorithms (DQN) in MinAtar.

\textbf{Building state representation} is fundamental for deep RL. The SF model is a type of \textit{general value function} \citep{Sutton2011HordeAS}, hypothesized to be a core component in building internal representations of autonomous agents \citep{Sutton2011HordeAS,White2015DEVELOPINGAP,Schlegel2018DiscoveryOP}. Our work ties to this topic, since we can view the partial SF model as a new \emph{learned} representation of the value function used as learning target in the \returnname.

\section{Discussion}
In this work we propose a new, generalized learning target that combines the previous approaches, making more efficient use of the same experience. The approach we proposed uses an implicit model represented by the SF model, and can thus also be viewed as implicit planning with a multi-step policy-dependent expectation model.
The \returnname\,we proposed for the learning target can easily be used in place of the bootstrap target used in any value-based algorithm (e.g. TD($n$), TD($\lambda$)), as we have illustrated in this work for one-step returns used by TD($0$). Empirically, we showed that this method, while using the same amount of sampled experience, is more effective, resulting in more efficient value function estimation and higher control performance.

Many potential directions of investigation have been left for future work.
(i) The \returnname\, contains a successor feature estimate, which could also be further leveraged for exploration and transfer.
(ii) \citet{Chelu2020ForethoughtAH} investigates the complementary properties of explicit forward and backward models and argues for the potential of optimally combining both ``forward'' and ``backward'' facing credit assignment schemes. Further, \citet{van2020expected} introduces expected eligibility traces as implicit backward models, a kind of ``predecessor features'' (time-reversed successor features). Future work can explore the differences and commonalities between implicit models in the forward and backward direction using our proposed SF model and expected eligibility traces. The right balance between using \textit{backward} credit assignment through the use of eligibility traces, and \textit{forward} prediction through predictive representations remains an open question with fundamental implications for learning efficiency.
(iii) How to best use predictive representations to build an internal agent state is central to generalization and efficient credit assignment. Our work opens up many exciting new questions for investigation in this direction.


\section*{Acknowledgments}
AGXC was supported by the NSERC CGS-M, FRQNT, and UNIQUE excellence scholarships. This work was supported by NSERC (Discovery Grant: RGPIN-2020-05105; Discovery Accelerator Supplement: RGPAS-2020-00031) and CIFAR (Canada AI Chair; Learning in Machine and Brains Fellowship) grants to BAR. This research was enabled in part by computational resources provided by Calcul Québec (\url{www.calculquebec.ca}) and Compute Canada (\url{www.computecanada.ca}). We thank the anonymous reviewers for their valuable feedback. We thank our colleagues at Mila for the insightful discussions that have made this project better: Emmanuel Bengio, Wesley Chung, Nishanth Anand, Maxime Wabartha, Harry Zhao, Andrei Lupu, David Tao, Etienne Denis, and Mandana Samiei.

\bibliography{references}

\onecolumn
\pdfoutput=1 

\begin{appendices}
\label{sec:appendix}

\section{Broader impact statement}
\label{sec:broader-impact}
Our results are theoretical and provide fundamental insights into reinforcement learning algorithms by proposing and investigating a spectrum of bootstrapped learning targets, including one-step TD (which directly estimates the next-state value) and the successor features value estimate (which separately estimates future cumulative features and instantaneous rewards to be linearly combined) as special cases. We hope our work will contribute to improved understanding of RL and the goal of developing generally intelligent real-world systems. However, we do not focus on applications in this work, and substantial additional work will be required to apply our methods to real-world settings. Regarding training resources, each run for the tabular experiments (figures \ref{fig:deterministic-chain-intuition} and \ref{fig:tabular-random-chain}) take $<1$ minute on CPU, and each Mini-Atari (figures \ref{fig:minatar-q-performance-overview} and \ref{fig:minatar-lr-sweep}) takes 7-10 hours on a single RTX8000 GPU on our internal cluster. We estimate the total training compute hours is 5.5-6k hours, for a carbon emission footprint of 673.92 kg CO2 as estimated by \url{https://mlco2.github.io/impact/}.

\section{Lambda return and the \returnname}
\label{Ap:lambda-return-equivalencies}

\subsection{Lambda return}

\begin{lemma}
    The infinite horizon discounted lambda return can be written equivalently in the following two ways:
    \begin{align}
        G_t^\lambda &= (1-\lambda) \sum_{n=1}^\infty \lambda^{n-1} \left[
            [ \sum_{k=1}^n \gamma^{k-1} R_{t+k} ] + \gamma^n v_\theta(S_{t\!+\!n})
        \right] \\
        &= R_{t+1} + \gamma \sum_{n=1}^{\infty} (\lambda\gamma)^{n-1} \left[ (1-\lambda) v_\theta(S_{t\!+\!n}) + \lambda R_{t+n+1} \right]
        \label{eq:lambda-return-sum-of-value-and-reward-form}
    \end{align}
\end{lemma}
\begin{proof}
We write $V_t = v_\theta(S_t) = \vect{\phi}(S_t)\!^\top \vtheta$ for brevity,
\begin{align}
    G_t^\lambda &= (1-\lambda) \sum_{n=1}^\infty \lambda^{n-1} \left[
        [ \sum_{k=1}^n \gamma^{k-1} R_{t+k} ] + \gamma^n V_{t+n}
    \right] \,,
    \\
    &= (1-\lambda)\lambda^0 [R_{t+1} + \gamma V_{t+1}] \nonumber\\
    &\quad + (1-\lambda)\lambda^1 [R_{t+1} + \gamma R_{t+2} + \gamma^2 V_{t+2}] \nonumber \\
    &\quad + ... \,,
    \label{eq:lambda-return-row-sums}
    \\
    &= R_{t+1} + (1-\lambda) \gamma V_{t+1} \nonumber\\
    &\quad + (\lambda\gamma) R_{t+2} + (1-\lambda) \lambda\gamma^2 V_{t+2} \nonumber \\
    &\quad + (\lambda\gamma)^2 R_{t+3} + (1-\lambda) \lambda^2\gamma^3 V_{t+3} \nonumber \\
    &\quad + ... \,,
    \label{eq:lambda-return-col-sums}
    \\
    &= (\lambda\gamma)^0 R_{t+1} + (\lambda\gamma)^0 [(1-\lambda)\gamma V_{t+1}] \nonumber \\
    &\quad + (\lambda\gamma)^1 R_{t+2} + (\lambda\gamma)^1 [(1-\lambda)\gamma V_{t+2}] \nonumber \\
    &\quad + (\lambda\gamma)^2 R_{t+3} + (\lambda\gamma)^2 [(1-\lambda)\gamma V_{t+3}] \nonumber \\
    &\quad + ... \,,
    \\
    &= R_{t+1} \nonumber + (\lambda\gamma)^0 [(1-\lambda)\gamma V_{t+1} + (\lambda\gamma) R_{t+2}] \nonumber \\
    &\hspace{35pt} + (\lambda\gamma)^1 [(1-\lambda)\gamma V_{t+2} + (\lambda\gamma) R_{t+3}] \nonumber \\
    &\hspace{35pt} + (\lambda\gamma)^2 [(1-\lambda)\gamma V_{t+3} + (\lambda\gamma) R_{t+4}] \nonumber \\
    &\hspace{35pt} + ... \,,
    \\
    &= R_{t+1} + \gamma \sum_{n=1}^{\infty} (\lambda\gamma)^{n-1} \left[ (1-\lambda) V_{t+n} + \lambda R_{t+n+1} \right] \,.
\end{align}
\end{proof}

We go from equation \eqref{eq:lambda-return-row-sums} to \eqref{eq:lambda-return-col-sums} by pulling out each column of rewards and noting that the $\lambda$'s sum to one, $(1-\lambda) \sum_{k=0}^\infty \lambda^k = 1$.

\subsection{The \returnname}

For notation purposes to avoid confusion, we henceforth write $\vflet$ in place of $\lambda$.
We first replace the sampled instantaneous reward in the $\lambda$-return (equation~\eqref{eq:lambda-return-sum-of-value-and-reward-form}) with an instantaneous reward function (equation~\eqref{eq:linear-reward-function}), $r_{\textbf{w}}(s) \approx \mathbb{E}_\pi [R_{t+1} | S_t = s]$,
\begin{align}
    G_t^{\vflet} \approx R_{t+1} + \gamma \sum_{n=1}^{\infty} ({\vflet}\gamma)^{n-1} \left[ (1-{\vflet}) v_\theta(S_{t\!+\!n}) + {\vflet} \, r_{\textbf{w}}(S_{t+n}) \right] \,.
    \label{eq:lambda-return-with-approx-reward-function}
\end{align}
The above equation~\eqref{eq:lambda-return-with-approx-reward-function} is equivalent to equation~\eqref{eq:eta-return-summation-def} in the main text. We now derive the \returnname, by putting the random variables after $t+1$ in expectation,
\begin{align}
    G_t^{\vflet} &= R_{t+1} + \gamma \,  \mathbb{E}_\pi\!\left[
        \sum_{n=1}^{\infty} (\vflet\gamma)^{n-1} [
            (1-\vflet)\, v_\theta(S_{t\!+\!n}) + \vflet\, r_{\vw}(S_{t\!+\!n})
        ]
    \right] \,,
    \\
    &= R_{t+1} + \gamma \, \mathbb{E}_\pi\!\left[
        \sum_{n=1}^{\infty} (\vflet\gamma)^{n-1} [
            (1-\vflet)\, \vphi_{t\!+\!n}\!^\top \vtheta + \vflet\, \vphi_{t\!+\!n}\!^\top \vw
        ]
    \right] \,,
    \\
    &= R_{t+1} + \gamma \, \mathbb{E}_\pi\!\left[
        \sum_{n=1}^{\infty} (\vflet\gamma)^{n-1} \vphi_{t\!+\!n}^\top
    \right] [(1-\vflet)\, \vtheta + \vflet\, \vw ] \,,
    \\
    &= R_{t+1} + \gamma \, \vpsi^{\vflet} (S_{t+1}) [(1-\vflet)\, \vtheta + \vflet\, \vw ] \,.
\end{align}
This gives us equation~\eqref{eq:eta-return-mixture-target-def} in the main text, which we arrive at by factorizing out the SF, $\vpsi^\vflet (S_{t+1}) = \mathbb{E}_\pi\![\sum_{n=0}^\infty (\vflet\gamma)^{n} \vphi_{t\!+\!1\!+\!n}]$, to be separately estimated.

\section{Proof of proposition \ref{prop: theta-fixed-point}}
\label{Ap:proof-of-fixed-point}

We consider the setting of policy evaluation with linear function approximation. Given a Markov Reward Process (MRP) $\mathcal{M}_\pi = \langle \mathcal{S}, r_\pi, P_\pi \rangle$ with state space $\mathcal{S}$, reward function $r_\pi (s) = \mathbb{E}_\pi [R_{t+1} | S_t = s]$, and policy dependent transition function $P_\pi(s'|s)$. Assuming there are finite countable number states, we can write the MRP in matrix form with transition matrix $\vn{P}_\pi \in \mathbb{R}^{|\mathcal{S}| \times |\mathcal{S}|}$, reward vector $\vn{R} \in \mathbb{R}^{|\mathcal{S}|}$, $\vn{R}_i = \mathbb{E}_\pi [R_{t+1} | S_t = i]$, along with discount factor $\gamma \in [0, 1)$. Let each state be described by a $d$-dimensional feature, and $\vPhi \in \mathbb{R}^{|\mathcal{S}| \times d}$ be the feature matrix. We assume $\vPhi$ have linearly independent columns.

We consider all learning to be done on-policy with one-step TD given single-step experience tuples $(S_t, R_{t+1}, S_{t+1})$. Let $\vn{D} \in \mathbb{R}^{|\mathcal{S}| \times |\mathcal{S}|}$ denote a diagonal matrix whose diagonal is the stationary distribution of the MRP.

The following sections are as follows: we first review pre-existing results on solving for the fixed-point of the regular linear TD(0) with direct value prediction, and the factorized value estimate using successor feature and an instantaneous reward model. Finally, we will present our main result in solving for the fixed point of on-policy learning with the one-step \returnname target.

\subsection{MF value fixed point}
\label{sec:mf-linear-td-fixed-point}

We first consider the traditional ``model-free'' value learning (we refer to this as the ``MF value''). Given a linear MF value function with parameter $\vtheta$,
\begin{equation}
    v_{\vtheta}(s_t) = \vphi_t^\top \vtheta \approx \mathbb{E}_\pi \left[
        \sum\nolimits_{n=0}^\infty \gamma^n R_{t+n+1} | S_t = s_t
    \right] \,.
\end{equation}

Given an experience tuple $(\vphi_t, R_{t+1}, \vphi_{t+1})$, doing TD(0) with the MF value uses the following update with step-size parameter $\alpha \in (0, 1]$,
\begin{equation}
    \vtheta_{t+1} = \vtheta_t + \alpha \left(
        R_{t+1} + \gamma \vphi_{t+1}^\top \vtheta_t - \vphi_t^\top \vtheta_t
    \right) \vphi_t \,.
\end{equation}

Linear TD belongs to a family of \emph{linear fixed-point} methods and solve for the following fixed point \citep{Parr2008AnAO},
\begin{equation}
    \vtheta = (\vPhi^\top \vn{D} \vPhi)^{-1} \vPhi^\top \vn{D} \left(
        \vn{R} + \gamma \vn{P}_\pi \vPhi \vtheta
    \right) \,.
\end{equation}

The above has the following fixed point solution, which is also referred to as the \emph{TD fixed point},
\begin{equation}
    \vtheta_{TD} = (\vPhi^\top \vn{D} \vPhi - \gamma \vPhi^\top \vn{D} \vn{P}_\pi \vPhi)^{-1} \vPhi^\top \vn{D} \vn{R} \,.
    \label{eq:linear-td-fixed-point}
\end{equation}

Note the fixed point implies the following,
\begin{equation}
    (\vPhi^\top \vn{D} \vPhi)^{-1} \vPhi^\top \vn{D} \left(
        \vn{R} + \gamma \vn{P}_\pi \vPhi \vtheta_{TD}
    \right) = \vtheta_{TD} \,.
    \label{eq:td-fixed-point-in-iterative-system}
\end{equation}

\paragraph{Discounting by $\vflet$} We can write a similar system with a $\vflet\gamma$-discounted value function, with the following fixed point,
\begin{align}
    & \vtheta_{TD}^\vflet = (\vPhi^\top \vn{D} \vPhi - \vflet \gamma \vPhi^\top \vn{D} \vn{P}_\pi \vPhi)^{-1} \vPhi^\top \vn{D} \vn{R} \,,
    \\
    & (\vPhi^\top \vn{D} \vPhi)^{-1} \vPhi^\top \vn{D} \left(
        \vn{R} + \vflet \gamma \vn{P}_\pi \vPhi \vtheta^\vflet{TD}
    \right) = \vtheta^\vflet_{TD} \,.
\end{align}

\subsection{SF value fixed point}
\label{sec:sf-linear-td-fixed-point}

Here we write the fixed point for a linear successor feature (SF) parameterized value function.

\subsubsection{SF fixed point}

Given linear successor features (SFs) with linear parameters ${\vZ} \in\mathbb{R}^{d \times d}$,
\begin{equation}
    \vpsi_{\vZ}(s) = {\vZ}^\top \vphi(s) \approx \mathbb{E}_\pi \left[
        \sum\nolimits _{n=0}^{\infty} {\gamma}^n \vphi_{t+n}
    \mid S_t=s \right]\,.
\end{equation}

With experience tuple $(\vphi_t, R_{t+1}, \vphi_{t+1})$, doing TD(0) for SF learning has the following update,
\begin{equation}
    \vZ_{t+1}^\top = \vZ_{t}^\top + \alpha \left(
        \vphi_t + \gamma \vZ_{t}^\top \vphi_{t+1} - \vZ_{t}^\top \vphi_{t}
    \right) \vphi_t^\top \,.
\end{equation}

Similar to value-learning with TD(0), SF learning with TD(0) corresponds to solving the following,
\begin{align}
    \vZ &= (\vPhi^\top \vn{D} \vPhi)^{-1} \vPhi^\top \vn{D} \left(
        \vPhi + \gamma \vn{P}_\pi \vPhi \vZ
    \right) \,,
    \\
    &= \vn{I} + \gamma (\vPhi^\top \vn{D} \vPhi)^{-1} \vPhi^\top \vn{D} \vn{P}_\pi \vPhi \vZ \,.
\end{align}

The SF fixed point is as follows,
\begin{equation}
    \vZ_{TD} = (\vPhi^\top \vn{D} \vPhi - \gamma \vPhi^\top \vn{D} \vn{P}_\pi \vPhi)^{-1} \vPhi^\top \vn{D} \vPhi \,.
\end{equation}

\paragraph{Discounting by $\vflet$}

We similarly denote a $\vflet\gamma$-discounted SF, $\vpsi^{\vflet}(s) = \mathbb{E}_\pi [ \sum\nolimits _{n=0}^{\infty} {(\vflet \gamma)}^n \vphi_{t+n} \mid S_t=s ]$, with the following fixed point,
\begin{align}
    & \vZ_{TD}^{\vflet} = (\vPhi^\top \vn{D} \vPhi - \vflet\gamma \vPhi^\top \vn{D} \vn{P}_\pi \vPhi)^{-1} \vPhi^\top \vn{D} \vPhi \,,
    \\
    & (\vPhi^\top \vn{D} \vPhi)^{-1} \vPhi^\top \vn{D} \left(
        \vPhi + \vflet \gamma \vn{P}_\pi \vPhi \vZ^{\vflet}_{TD}
    \right) = \vZ^{\vflet}_{TD}\,.
\end{align}

\subsubsection{Reward regression solution}

Given a linear reward function with parameters $\vw$, estimating the instantaneous reward,
\begin{equation}
    \vn{r}_{\vw} (s_t) = \phi_t^\top \vw
    \approx \mathbb{E}_\pi [R_{t+1} | S_t = s_t ] \,.
\end{equation}

With experience tuple $(\vphi_t, R_{t+1}, \vphi_{t+1})$, reward learning follows a supervised update,
\begin{equation}
    \vw_{t+1} = \vw_t + \alpha \left(
        R_{t+1} - \vphi_t^\top \vw_t
    \right) \vphi_t \,.
\end{equation}

The reward regression solution is,
\begin{align}
    \hat{\vw} = (\vPhi^\top \vn{D} \vPhi)^{-1} \vPhi^\top \vn{D} \vn{R} \,.
\end{align}

\subsubsection{SF Value}

We construct the SF value estimate as a dot product $v_{\vpsi} (s) = \vpsi_\pi (s)\!^\top\! \cdot \vw$, written in matrix form we get,
\begin{align}
    \vn{v} &= \vn{\Psi}_{\vZ} \cdot \vw \,,
    \\
    &= \vPhi \vZ \vw \,.
\end{align}

At the parameters' respective fixed points, we recover the value (TD) fixed point (equation~\ref{eq:linear-td-fixed-point}),
\begin{align}
    \vZ_{TD} \hat{\vw} &= (\vPhi^\top \vn{D} \vPhi - \gamma \vPhi^\top \vn{D} \vn{P}_\pi \vPhi)^{-1} (\vPhi^\top \vn{D} \vPhi) \, (\vPhi^\top \vn{D} \vPhi)^{-1} \vPhi^\top \vn{D} \vn{R} \,,
    \\
    &= (\vPhi^\top \vn{D} \vPhi - \gamma \vPhi^\top \vn{D} \vn{P}_\pi \vPhi)^{-1} \vPhi^\top \vn{D} \vn{R} \, = \vtheta_{TD} \,.
\end{align}

Similarly, we have $\vZ_{TD}^{\vflet} \hat{\vw} = \vtheta_{TD}^\vflet$.

\subsection{Linear \returnname\, value fixed point}
\label{sec:lvf-linear-td-fixed-point}

We now consider doing value learning with the \returnname. Given an experience tuple $(\vphi_t, R_{t+1}, \vphi_{t+1})$, hyper-parameter $\vflet \in [0,1]$, and some SF and reward parameters $\vZ, \vw$, the one-step \returnname has the following update,
\begin{align}
    \vtheta_{t+1} &= \vtheta_t + \alpha \left(
        R_{t+1} + \gamma \, (\vpsi_{t+1}^\vflet)^\top [(1-\vflet) \vtheta_t + \vflet \vw] - \vphi_t^\top \vtheta_t
    \right) \vphi_t \,,
    \\
    &= \vtheta_t + \alpha \left(
        R_{t+1} + \gamma \, \vphi_{t+1}^\top \vZ [(1-\vflet) \vtheta_t + \vflet \vw] - \vphi_t^\top \vtheta_t
    \right) \vphi_t \,.
\end{align}

Written in matrix form, the above iteration solves for the following,
\begin{align}
    \vtheta &= (\vPhi^\top \vn{D} \vPhi)^{-1} \vPhi^\top \vn{D} \left(
        \vn{R} + \gamma [(1-\vflet) \vn{P}_\pi \vPhi \vZ \vtheta + \vflet \vn{P}_\pi \vPhi \vZ \vw ]
    \right) \,,
    \\
    &= (\vPhi^\top \vn{D} \vPhi)^{-1} \vPhi^\top \vn{D} \left(
        \vn{R} + \gamma \vn{P}_\pi \vPhi [(1-\vflet) \vZ \vtheta + \vflet \vZ \vw ]
    \right) \,.
    \label{eq:linear-lvf-td0-projected-system}
\end{align}

We show the above system has the \emph{TD fixed point}, $\vtheta_{TD}$, as its fixed point as well.

\begin{lemma}
    Assuming the inverse of the SF parameters matrix $\vZ^{\vflet}_{TD}$ exists,\footnote{This condition holds given the above assumption of $\vPhi$ having linearly independent columns.} $(\vZ^{\vflet}_{TD})^{-1} = (\vPhi^\top \vn{D} \vPhi)^{-1} (\vPhi^\top \vn{D} \vPhi - \vflet\gamma \vPhi^\top \vn{D} \vn{P}_\pi \vPhi)$, the following identity holds,
    \begin{equation}
        (\vZ^\vflet_{TD})^{-1} \vtheta_{TD} = (1-\vflet) \vtheta_{TD} + \vflet (\vZ^\vflet_{TD})^{-1} \vtheta^\vflet_{TD} \,.
    \end{equation}
    \label{lemma:theta-lambda-theta-fp-relationship}
\end{lemma}
\begin{proof}
    The above identity can be re-arranged to the following form,
    \begin{align}
        \vflet (\vZ^{\vflet}_{TD})^{-1} \vtheta^\vflet_{TD} = (\vZ^{\vflet}_{TD})^{-1} \vtheta_{TD} - (1-\vflet) \vtheta_{TD} \,.
    \end{align}
    We show the left- and right-hand sides are equivalent. We first evaluate the right hand side (r.h.s.), substituting in the fixed point equations,
    \begin{align}
        & (\vZ^{\vflet}_{TD})^{-1} \cdot \vtheta_{TD} - (1-\vflet) \vtheta_{TD} \,, \nonumber
        \\
        =& \left( (\vZ^{\vflet}_{TD})^{-1} - (1-\vflet) \vn{I} \right) \, \vtheta_{TD} \,,
        \\
        =& \left( (\vPhi^\top \vn{D} \vPhi)^{-1} (\vPhi^\top \vn{D} \vPhi - \vflet\gamma \vPhi^\top \vn{D} \vn{P}_\pi \vPhi) - \vn{I} + \vflet \vn{I} \right) \, \vtheta_{TD} \,,
        \\
        =& \left( \vn{I} - \vflet\gamma (\vPhi^\top \vn{D} \vPhi)^{-1} \vPhi^\top \vn{D} \vn{P}_\pi \vPhi - \vn{I} + \vflet \vn{I} \right) \, \vtheta_{TD} \,,
        \\
        =& \vflet  \left( \vn{I} - \gamma (\vPhi^\top \vn{D} \vPhi)^{-1} \vPhi^\top \vn{D} \vn{P}_\pi \vPhi \right) \, \vtheta_{TD} \,,
        \\
        =& \vflet \, (\vPhi^\top \vn{D} \vPhi)^{-1} (
            \vPhi^\top \vn{D} \vPhi - \gamma \vPhi^\top \vn{D} \vn{P}_\pi \vPhi
        ) \, \vtheta_{TD} \,,
        \\
        =& \vflet \, (\vPhi^\top \vn{D} \vPhi)^{-1} (
            \vPhi^\top \vn{D} \vPhi - \gamma \vPhi^\top \vn{D} \vn{P}_\pi \vPhi
        ) \, (\vPhi^\top \vn{D} \vPhi - \gamma \vPhi^\top \vn{D} \vn{P}_\pi \vPhi)^{-1} \vPhi^\top \vn{D} \vn{R} \,,
        \\
        =& \vflet \, (\vPhi^\top \vn{D} \vPhi)^{-1} \vPhi^\top \vn{D} \vn{R} \,.
    \end{align}
    We evaluate the left hand side (l.h.s.),
    \begin{align}
        & \vflet (\vZ^{\vflet}_{TD})^{-1} \vtheta^\vflet_{TD} \nonumber
        \\
        =& \vflet (\vPhi^\top \vn{D} \vPhi)^{-1} (\vPhi^\top \vn{D} \vPhi - \vflet\gamma \vPhi^\top \vn{D} \vn{P}_\pi \vPhi) \, (\vPhi^\top \vn{D} \vPhi - \vflet \gamma \vPhi^\top \vn{D} \vn{P}_\pi \vPhi)^{-1} \vPhi^\top \vn{D} \vn{R} \,,
        \\
        =& \vflet (\vPhi^\top \vn{D} \vPhi)^{-1} \vPhi^\top \vn{D} \vn{R} \,.
    \end{align}
    The l.h.s. and r.h.s. are the same. It should be noted that all steps are invertible.
\end{proof}

\begin{proposition}
    \textbf{(Proof of Proposition \ref{prop: theta-fixed-point} in main text}) Given the SF parameter is at its $\vflet\gamma$-discounted fixed point, $\vZ^{\vflet}_{TD}$, and the reward parameter is at its supervised regression solution, $\hat{\vw}$, on-policy one-step learning with the \returnname\, has the TD fixed point, $\vtheta_{TD}$ as its fixed point,
    \begin{equation}
        (\vPhi^\top \vn{D} \vPhi)^{-1} \vPhi^\top \vn{D} \left(
            \vn{R} + \gamma \vn{P}_\pi \vPhi [(1-\vflet) \vZ^\vflet_{TD} \vtheta_{TD} + \vflet \vZ^\vflet_{TD} \hat{\vw} ]
        \right) = \vtheta_{TD} \,.
    \end{equation}
\end{proposition}
\begin{proof}
    We first substitute the fixed points $\vZ_{TD}^{\vflet}$ and $\hat{\vw}$ into the system (equation~\ref{eq:linear-lvf-td0-projected-system}), and use the identity $\vZ_{TD}^{\vflet} \hat{\vw} = \vtheta_{TD}^\vflet$,
    \begin{align}
        \vtheta_{t+1} &= (\vPhi^\top \vn{D} \vPhi)^{-1} \vPhi^\top \vn{D} \left(
            \vn{R} + \gamma \vn{P}_\pi \vPhi [(1-\vflet) \vZ^{\vflet}_{TD} \vtheta_t + \vflet \vZ^{\vflet}_{TD} \hat{\vw} ]
        \right) \,,
        \\
        &= (\vPhi^\top \vn{D} \vPhi)^{-1} \vPhi^\top \vn{D} \left(
            \vn{R} + \gamma \vn{P}_\pi \vPhi [(1-\vflet) \vZ^{\vflet}_{TD} \vtheta_t + \vflet \vtheta^{\vflet}_{TD} ]
        \right) \,.
    \end{align}

    We verify that $\vtheta_{TD}$ is the fixed-point by considering the case where $\vtheta_t = \vtheta_{TD}$,
    \begin{align}
        \vtheta_{t+1} &= (\vPhi^\top \vn{D} \vPhi)^{-1} \vPhi^\top \vn{D} \left(
            \vn{R} + \gamma \vn{P}_\pi \vPhi [(1-\vflet) \vZ^{\vflet}_{TD} \vtheta_{TD} + \vflet \vtheta^{\vflet}_{TD} ]
        \right) \,,
        \\
        &= (\vPhi^\top \vn{D} \vPhi)^{-1} \vPhi^\top \vn{D} \left(
            \vn{R} + \gamma \vn{P}_\pi \vPhi \, \vZ^{\vflet}_{TD} \, [(1-\vflet) \vtheta_{TD} + \vflet (\vZ^{\vflet}_{TD})^{-1} \vtheta^{\vflet}_{TD} ]
        \right) \,,
        \\
        &= (\vPhi^\top \vn{D} \vPhi)^{-1} \vPhi^\top \vn{D} (
            \vn{R} + \gamma \vn{P}_\pi \vPhi \, \vZ^{\vflet}_{TD} (\vZ^{\vflet}_{TD})^{-1} \, \vtheta_{TD}
        ) \,, &\text{[Lemma \ref{lemma:theta-lambda-theta-fp-relationship}]}
        \\
        &= (\vPhi^\top \vn{D} \vPhi)^{-1} \vPhi^\top \vn{D} (
            \vn{R} + \gamma \vn{P}_\pi \vPhi \vtheta_{TD}
        ) \,,
        \\
        &= \vtheta_{TD} \,. &\text{[By Eq.\ref{eq:td-fixed-point-in-iterative-system}]}
    \end{align}
    We see $\vtheta_t = \vtheta_{t+1} = \vtheta_{TD}$, therefore $\vtheta_{TD}$ is the fixed point of the system.
\end{proof}

In the main text proposition \ref{prop: theta-fixed-point} we have written the above fixed point in an equivalent expectation form for clarity, and used $\vtheta_{TD} = \vtheta_{TD(0)}$ in order to emphasize the point that this is the same fixed point as linear TD(0) (i.e. using the one-step TD return as a learning target).

\section{Experimental Set-up}
\label{Ap:experiment-details}

\subsection{Value prediction in a deterministic chain}

We set up the 16 state deterministic chain. As everything is deterministic, we set the learning rate $\alpha=1.0$ so new information can be learned right away. Thus the main point here is to see the speed of best possible (one-step transition-based) credit propagation.

\subsection{Value prediction in a random chain}

We compare the algorithms in a prediction setting in the 19-state random walk chain with tabular features. We train in the \textit{online incremental} setting---the agent receives a stream of episodic experiences $(S_1, R_1, S_2, R_2, ...)$, and updates its parameter immediately upon receiving the most recent one-step experience tuple (for example, $(S_{t-1}, R_t, S_t)$ at timestep $t$). Table \ref{table:random-walk-chain-experiment-parameters} details the parameters tested.

\begin{table}[]
  \centering
  \begin{tabular}{ll}
    \toprule
    Parameters  & Parameter values \\
    \midrule
    Value parameters learning rate, $\alpha_{\vtheta}$ & Sweep over $\{0.01, 0.1, 0.2, 0.3, 0.5\}$ \\
    Successor features learning rate, $\alpha_{\vZ}$ & Always same as $\alpha_{\vtheta}$  \\
    Reward learning rate, $\alpha_{\vw}$ & Always same as $\alpha_{\vtheta}$ \\
    $\vflet$ of the \returnname\, & Sweep over $\{ 0.0, 0.3, 0.5, 0.7, 0.9, 0.99, 1.0 \}$ \\
    Random seeds & $\{ 2,4,6,8,10,12,14,16,18,20 \}$ \\
    \bottomrule
  \end{tabular}
  \vspace{5pt}
  \caption{Experimental parameters of random walk chain}
  \label{table:random-walk-chain-experiment-parameters}
\end{table}

\subsection{Nonlinear architecture}
\label{Ap:lambda-dqn}

\begin{figure}
    \centering

    \textbf{(A)}
    \subfloat{\scalebox{0.19}{
        \includegraphics{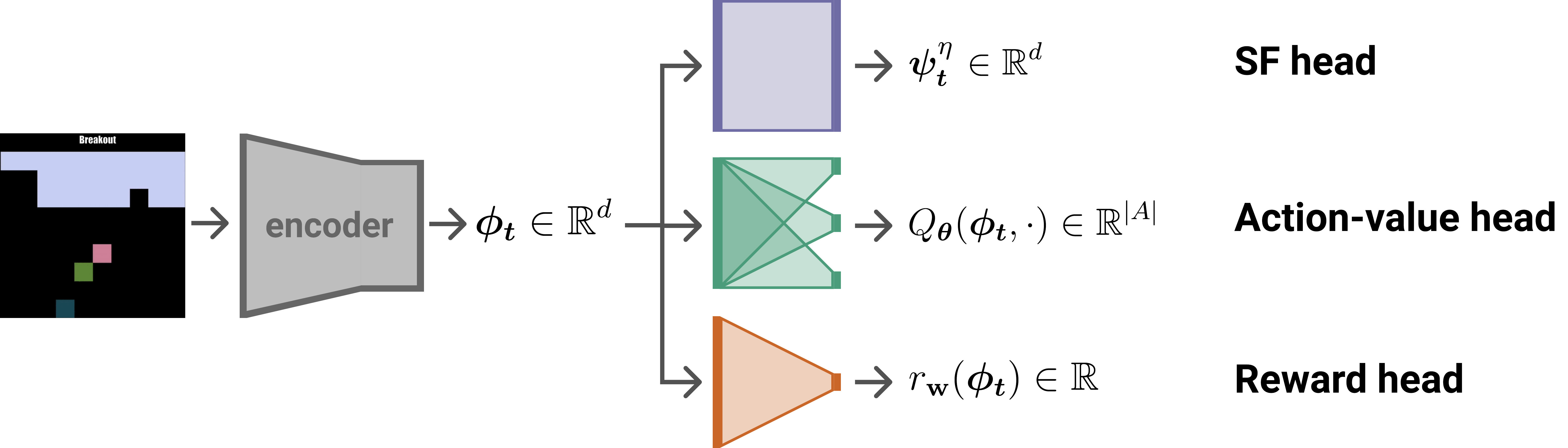}
        }} \vspace{10pt}

    \textbf{(B)}
    \subfloat{\scalebox{0.19}{
        \includegraphics{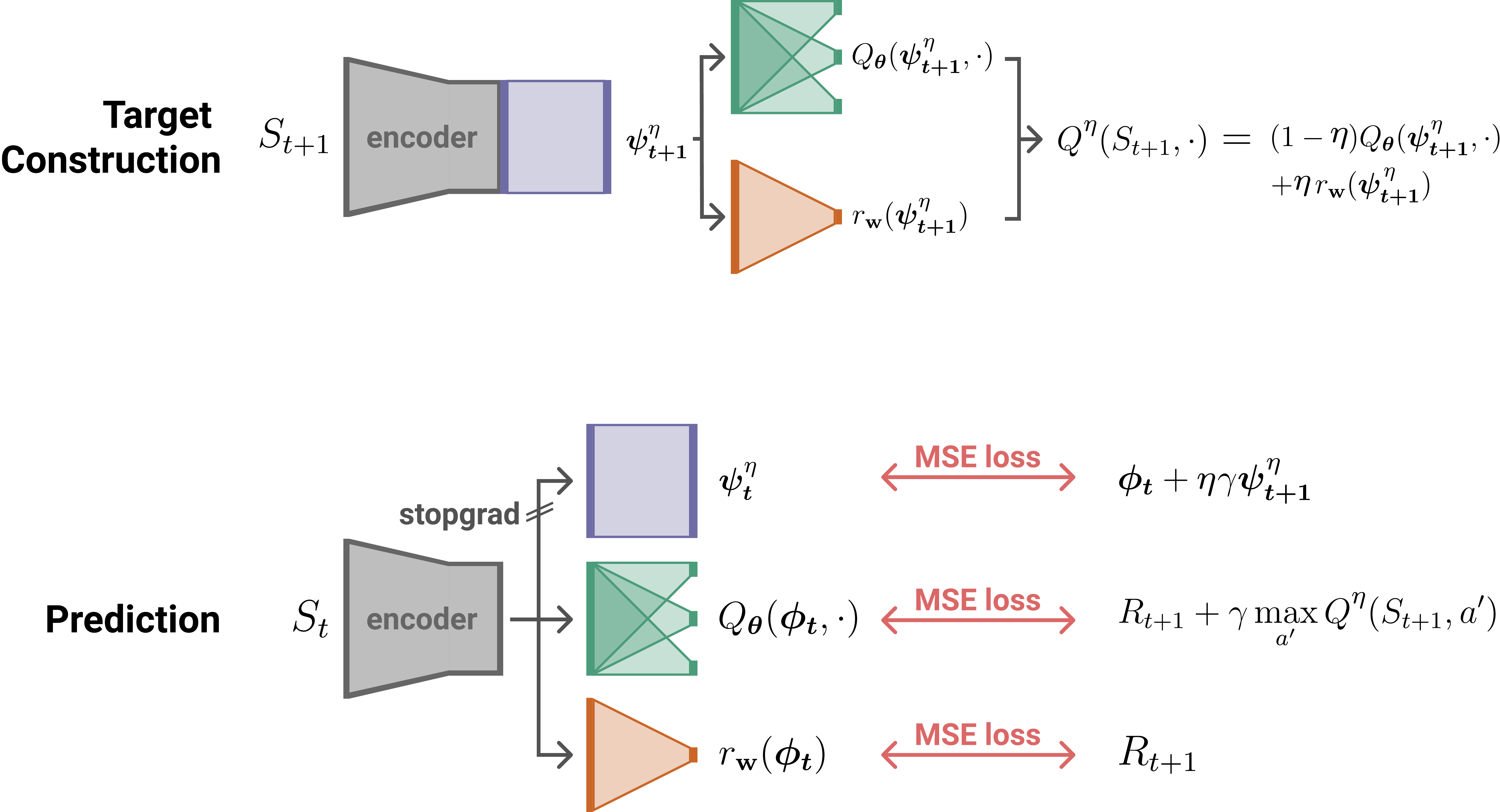}
        }}

    \caption{\textbf{Architecture for the \returnname\, augmented Deep Q Network}. \textbf{(A) Base architecture}, we augment a DQN-like architecture \citep{mnih2015human} (encoder and action-value head) with two additional heads for SF prediction and instantaneous reward prediction. \textbf{(B) Training with the \returnname\,}, given an experience tuple $(S_t, A_t, R_{t+1}, S_{t+1})$, we use $S_{t+1}$ to generate the next step estimate $Q^\vflet (S_{t+1}, \cdot)$ in constructing the target, and $S_t$ to generate the current step predictions. Training is done by minimizing the (MSE) loss between the prediction and targets.}
    \label{fig:lambda-dqn-architecture}
\end{figure}

Figure \ref{fig:lambda-dqn-architecture} details the architecture design of a deep Q network (based on the DQN architecture of \citep{mnih2015human}) trained using an \returnname\, to do nonlinear action-value function approximation. The corresponding pseudo-code is detailed in algorithm \ref{alg:nonlinear-lambda-fitted-q}.

\begin{algorithm}
    \caption{Deep Fitted Q Iteration with \returnname}
    \label{alg:nonlinear-lambda-fitted-q}
    \textbf{Input}:
        Feature encoder $\vphi_\xi(s_t) = \vphi_t$, where $\vphi_t \in \mathbb{R}^d$, with nonlinear parameters $\xi$, \\
        {\null\hskip2.6em\relax} Action value layer $q_{\vtheta} ({\vphi}_t, a) = {\vphi}_t^\top {\vtheta}_a$, where $q_{\vtheta} : \mathbb{R}^d \rightarrow \mathbb{R}^{|A|}$, with linear parameters ${\vtheta}$, \\
        {\null\hskip2.6em\relax} Reward layer $\text{r}_{\vw}(\vphi_t) = {\vphi}_t^\top {\vw}$, where $r_{\vw}: \mathbb{R}^d \rightarrow \mathbb{R}$, with linear parameters ${\vw}$, \\
        {\null\hskip2.6em\relax} SF layer ${\vpsi}_{\vZ}^\vflet(\vphi_t) = {\vZ}^\top \vphi_t$, where ${\vpsi}_{\vZ}^\vflet : \mathbb{R}^d \rightarrow \mathbb{R}^d$, with linear parameters $
        {\vZ}$, \\
        {\null\hskip2.6em\relax} Hyper-parameters $\gamma \in [0, 1)$, $\vflet \in [0,1]$ .\\
    \textbf{Output}: Deep Q function $q_{\vtheta} (\vphi_t, a)$, with encoder $\vphi_t \leftarrow \vphi_\xi(s_t)$, for control. \\
    \vspace{-1em} 
    \begin{algorithmic}[1] 
        \FOR{each environment step}
            \STATE Sample experience $(s_t, a_t, r_{t+1}, s_{t+1})$ from environment,

            \STATE Store to buffer $\mathcal{B} = \{ \mathcal{B} \cup (s_t, a_t, r_{t+1}, s_{t+1}) \}$ ,

            \STATE Sample i.i.d. minibatch of size $n$ from buffer $\{(s_k, a_k, r_{k+1}, s_{k+1})_{i=1, ..., n}\} \sim \mathcal{B}$ , \Comment{For Fitted Q Iteration}

            \FOR{each minibatch tuple $(s_k, a_k, r_{k+1}, s_{k+1})_i$}
                \STATE Encode features: $\vphi_k$, $\vphi_{k+1}$ $\leftarrow$ $\vphi_\xi(s_k)$, $\vphi_\xi(s_{k+1})$,

                \STATE Compute successor features: $\vpsi_{k+1} \leftarrow \vpsi_{\vZ}^\vflet (\vphi_{k+1})$,

                \STATE Copy feature with stop gradient (\texttt{sg}): $\vphi^{de}_k$ $\leftarrow$ $(\vphi_k)\texttt{.sg()}$

                \STATE $\mathcal{L}_{S,i} = \nicefrac{1}{2} \left[
                    [\vphi_k^{de} + \vflet\gamma \vpsi_{k+1}]\texttt{.sg()} - \vpsi_{\vZ}^\vflet(\vphi_k^{de})
                \right]^2$ ,
                \Comment{Successor features TD learning loss}

                \STATE $\mathcal{L}_{R,i} = \nicefrac{1}{2} \left[
                    r_{k+1} - \text{r}_{\vw}(\vphi_k)
                \right]^2$ ,
                \Comment{Reward supervised learning loss}

                \STATE $q^\vflet (s_{k+1}, a') = (1-\vflet) q_\theta (\vpsi_{k+1}, a') + \vflet \text{r}_{\vw}(\vpsi_{k+1})$ ,
                \Comment{Construct the \returnname}

                \STATE $\mathcal{L}_{Q,i} = \nicefrac{1}{2} \left[
                    [r_{k+1} + \gamma \max_{a'} q^\vflet (s_{k+1}, a')]\texttt{.sg()} - q_{\vtheta} (s_k, a_k)
                \right]^2$ .
                \Comment{Q learning loss}
            \ENDFOR

            \STATE$\mathcal{L}_{total} = \frac{1}{n} \sum_{i=1, ..., n} \mathcal{L}_{S,i} + \mathcal{L}_{R,i} + \mathcal{L}_{Q,i}$ ,
            \Comment{Overall minibatch loss}

            \STATE $\xi, \vtheta, \vw, \vZ$ $\leftarrow$ $\texttt{Optimizer}(\mathcal{L}_{total})$ .
            \Comment{Backprop and update parameters}

        \ENDFOR
    \end{algorithmic}
    \label{alg:deep-lambda-q-learning}
\end{algorithm}

\subsection{Mini-Atari: experiment section \ref{sec:minatar-control}}
\label{Ap:minatar-experiment-details}

We use the MinAtar environment \citep{young2019minatar}.\footnote{GitHub commit: \url{https://github.com/kenjyoung/MinAtar/tree/8fceb584a00d86a3294c2d6ffb6fb8d93496b6a5}} We build our ``deep $\vflet$-Q'' agent based on the DQN provided by \citet{young2019minatar} in \texttt{examples/dqn.py}. We implement the same DQN architecture and replicate to the best of our abilities the same hyperparameters as \citet{young2019minatar}, which was built to mimic the architecture and training procedure of the original DQN of \citet{mnih2015human}, albeit miniaturized for the smaller Atari environments. Unlike the original DQN, training is done every frame, using the PyTorch \citep{torch2019} implementation of the RMSprop optimizer \citep{tieleman2012rmsprop}. We report in details the architectures in figure~\ref{fig:lambda-dqn-architecture} and the training hyperparameters in table \ref{table:default-minatar-params}.

\begin{table}
  \centering
  \begin{tabular}{ll}
    \toprule
    Hyperparameter  & Value \\
    \midrule
    discount factor ($\gamma$) & 0.99 \\
    replay buffer memory size & 100000 \\
    replay sampling minibatch size & 32 \\
    $\epsilon$-greedy policy, initial $\epsilon$ & 1.0 \\
    Initial random exploration & 5000 environment steps \\
    $\epsilon$-greedy policy, final $\epsilon$ & 0.1 \\
    Initial to final $\epsilon$ anneal period & 100000 environment steps \\
    Target update period (per $n$ policy net updates) & 1000 \\
    RMSprop momentum & 0.0 \\
    RMSprop smoothing constant (alpha) & 0.95 \\
    RMSprop eps (added to denominator for num stability) & 0.01 \\
    RMSprop centered (normalized gradient) & True \\
    Learning rate (conv torso and value head, $\alpha_{\vtheta}$) & 0.00025 \\
    Learning rate (SF head, $\alpha_{\vZ}$) & 0.005 \\
    Learning rate (reward head, $\alpha_{\vw}$) & 0.005 \\
    \bottomrule
  \end{tabular}
  \vspace{5pt}
  \caption{Default hyperparameters for DQN and deep $\vflet$-Q network}
  \label{table:default-minatar-params}
\end{table}

Specifically, the main text results presented in figure~\ref{fig:minatar-q-performance-overview} follows exactly the hyperparameters reported in table \ref{table:default-minatar-params}, along with evaluations for a number of $\vflet$'s for a parameter study of $\vflet = \{0.0, 0.4, 0.5, 0.7, 0.95, 1.0\}$. Each setting was conducted for 10 independent runs with seeds $\{{2, 5, 8, 11, 14, 17, 20, 23, 26, 29}\}$.

Main text figure~\ref{fig:minatar-lr-sweep} conducts a parameter study on the learning rates of the individual components of the deep $\vflet$-Q agent: the value head and convolutional torso ($\alpha_{\vtheta}$, these two components make up exactly the ``vanilla'' DQN), the successor feature head ($\alpha_{\vZ}$), and the reward prediction head ($\alpha_{\vw}$). We investigated learning rates $\{0.00025, 0.0005, 0.001, 0.0025, 0.005\}$, and also compare our deep $\vflet$-Q agent (with intermediate $\vflet = 0.4$) against a ``vanilla'' DQN agent.

Averaging of return is done during training by averaging over the episodic return (total undiscounted reward received in a single episode) of 10 episodes. The steps are ``binned'' into increments of length $1e4$ to account for the fact that different runs will generate episodic returns at different environmental steps, making it difficult to compute confidence interval in a ``per-step'' way. That is, the logged steps (x-axis of training plots) are rounded to the nearest multiple of $1e4$ for all runs.

\newpage

\section{Additional results}
\label{Ap:additional-results}

\subsection{Measuring representation collapse}

For deep Q learning, we learn the feature representation $\vphi(\cdot)$ simultaneously to the successor features, action-values, and rewards (which are based on the learned feature layer). As our feature representation is shaped by back-propagated gradients from the action-value and reward heads (see figure~\ref{fig:lambda-dqn-architecture} and algorithm \ref{alg:nonlinear-lambda-fitted-q}), we measure the informativeness of the learned representation for different values of $\vflet$ (of the expected $\vflet$-return). Concretely, we measure the \emph{effective rank} \citep{yang2019harnessing,kumar2020implicit} of the feature learned after $5e6$ training environment steps, measured as,
\begin{equation}
    \texttt{srank}(\Phi) = \min \left\{k: \frac{\textstyle\sum_{i=1}^{k} \sigma_i (\Phi) }{\textstyle\sum_{i=1}^{d} \sigma_i (\Phi)} \geq 1 - \delta \right\} \,,
\end{equation}
with $\sigma_i(\Phi)$ being the $i$-th singular value of matrix $\Phi$, in decreasing order. We set $\delta = 0.01$ similar to \citet{kumar2020implicit}. Since we do not have access to the full feature matrix for MinAtar, we approximate $\Phi$ by sampling a large ($n=2048$) minibatch of samples from the replay buffer and encoding them using the convolutions torso for a matrix $\hat{\Phi} \in \mathbb{R}^{n \times d}$, $n=2048$, $d=128$. We measure the averaged \texttt{srank} for 8 sampled minibatches per run, though standard deviation is low between the independently sampled minibatches.

\begin{figure}
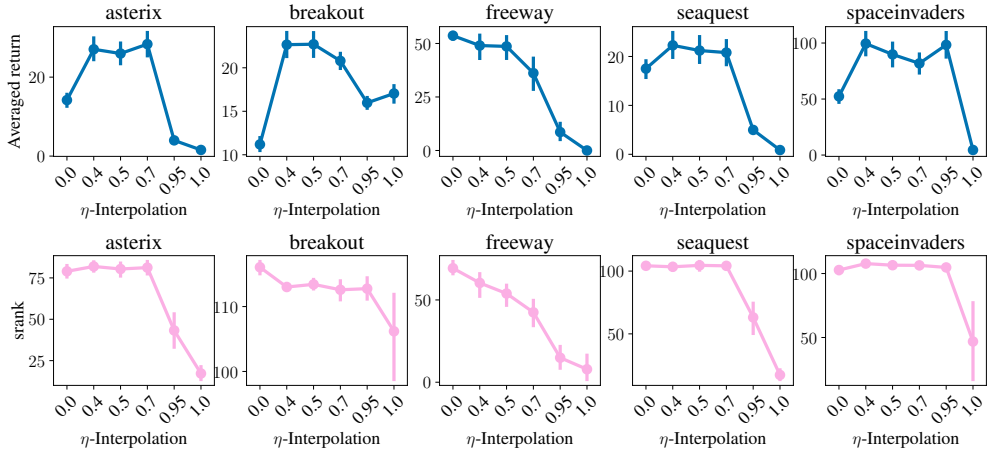

    \centering
    \subfloat{\scalebox{0.45}{
        \input{figs/minatar/minatar_eval_lambdas.pgf}
    }} \vspace{-1.2em}

    \subfloat{\scalebox{0.45}{
        \input{figs/minatar/minatar_eval_srank.pgf}
    }}

    \caption{\textbf{MinAtar evaluation performance and \textit{srank}}. \textbf{(Top)} Parameter study for different values of $\vflet$. Identical to figure \ref{fig:minatar-q-performance-overview}-A. \textbf{(Bottom) SRank}. The y-axis displays the average \texttt{srank} over $10$ independent seeds. Shaded area and error bars depicts $95$ confidence interval.}
    \label{fig:srank-eval}
\end{figure}

Figure~\ref{fig:srank-eval} (bottom) reports the srank for the same models as figure~\ref{fig:minatar-q-performance-overview} (we duplicate figure~\ref{fig:minatar-q-performance-overview}-A in the top row here). The maximum possible srank achievable is $128$ (i.e. the feature dimension, $d$), with lower srank indicating the learned feature representation is less informative. We observe that in general, srank is high ($>75$) and similar for $\vflet$'s up to $\vflet = 0.7$ (with the exception of \texttt{freeway}, to be discussed later). However, for high $\vflet$'s ($\vflet = \{0.95, 1.0\}$), we observe a decrease in srank for \nicefrac{4}{5} MinAtar games, with $\vflet=1.0$ having srank's that tend towards 0. In the case of \texttt{freeway}, srank decreases monotically as we increase $\vflet$. We hypothesize this is the result of \emph{sparse reward} for \texttt{freeway}. Since the feature layer is shaped in part by reward gradients, sparse reward may push the features to be less informative---an issue that is worsened as we depend more on the feature prediction rather than value prediction with higher $\vflet$'s.

Importantly, we observe the evaluation performance is related to the feature srank. Specifically, in cases where feature srank is similar, an intermediate $\vflet$ value out-performs ``extreme'' values of $\vflet$ (e.g. $\vflet = 0$). However, higher $\vflet$ appear to suffer from representation collapse which worsen performance, especially in sparse reward settings. The issue of learning good representation for successor feature learning can be addressed using auxiliary objectives (such as image reconstruction in \citet{kulkarni2016deepsuc} or next-state prediction in \citet{machado2020count}). We leave the interplay between the \returnname\, and additional feature-learning auxiliary tasks for future investigation.


\newpage

\bibliographystyle{apalike}
\bibliography{references}

\end{appendices}

\end{document}